\documentclass[journal]{IEEEtran}
\ifCLASSINFOpdf
\else
\fi

\usepackage[dvips]{graphicx}
\usepackage{multirow}
\usepackage{amsmath}
\usepackage{dsfont}
\usepackage{tabularx}
\usepackage{comment}
\usepackage[hidelinks]{hyperref}

\usepackage{pstool}
\usepackage{mathtools}
\usepackage{color}
\usepackage{psfrag}
\usepackage{caption}
\usepackage{pdfpages}
\usepackage{soul}

\usepackage{pifont}
\usepackage{url}

\newcommand{\boldphi}{\boldsymbol{\varphi}}
\newcommand{\bgamma}{\boldsymbol{\gamma}}
    \usepackage{latexsym}
\usepackage{subfigure}
\usepackage{amssymb}
\usepackage{url}
\usepackage{amsmath}
\usepackage{amssymb}
\usepackage{psfrag}
\usepackage[section]{placeins}
\usepackage{arydshln}

\def\atan2{{\mbox{atan2}}}

\newcommand{\Fig}[1]{Fig.~\ref{fig:#1}}

\def\12{\frac{1}{2}}

  \def\b{\mbox{\boldmath $b$}}
 
  \def\g{\mbox{\boldmath $g$}} 
 \def\j{\mbox{\boldmath $j$}}  
   \def\p{\mbox{\boldmath $p$}}
\def\q{\mbox{\boldmath $q$}}   
   \def\x{\mbox{\boldmath $x$}}
\def\y{\mbox{\boldmath $y$}} 

  \def\C{\mbox{\boldmath $C$}} 
 \def\F{\mbox{\boldmath $F$}}  
\def\I{\mbox{\boldmath $I$}} \def\J{\mbox{\boldmath $J$}} \def\K{\mbox{\boldmath $K$}} 
\def\M{\mbox{\boldmath $M$}}

\def\0{{\bf 0}}

 \def\b0{{\mbox{\boldmath $0$}}}

\def\natural{\mbox{\rm I\kern-0.2em N}}  % Symbol for the set of real numbers.

\def\real{\mbox{\rm I\kern-0.2em R}}  % Symbol for the set of real numbers.
\def\=def{\stackrel{def}{=}}

\def\beq{\begin{equation}}
\def\eeq{\end{equation}}
\def\be{\begin{equation}}
\def\ee{\end{equation}}
\def\bea{\begin{eqnarray}}
\def\eea{\end{eqnarray}}
\def\beann{\begin{eqnarray*}}
\def\eeann{\end{eqnarray*}}

\def\ds{\displaystyle}

  {\setlength{\arraycolsep}{0.14em} %% 0.14em = 2.5/18 quad = 0.5 thickspace,
   \begin{eqnarray}}%%%                default value was 5 pt (much bigger)
  {\end{eqnarray}}           %% see The TeXbook, p. 167-170

\def\Beginboxit
   {\par
    \vbox\bgroup
           \hrule height1pt
           \hbox\bgroup
                  \vrule width1pt \kern2pt %
                  \vbox\bgroup\kern2pt
   }
\def\Endboxit{%
                             \kern2pt
                             \vspace*{4pt}
                       \egroup
                  \kern2pt\vrule width1pt
                \egroup
           \hrule height1pt
         \egroup
   }

\newenvironment{boxit}{\Beginboxit}{\Endboxit}
\newenvironment{boxit*}{\Beginboxit\hbox to\hsize{}}{\Endboxit}

\def\ds{\displaystyle}

 \newif\ifSmallFigure \SmallFiguretrue
 \ifSmallFigure
    \def\FIGT2{c:/usr/claudio/work/ucima/tlibro/tfigure1} %tanto non lo usa
 \else
    \def\FIGT2{c:/usr/claudio/work/ucima/tlibro/tfigure} %tanto non lo usa
 \fi

\def\dotq{\dot{q}}
\def\ddotq{\ddot{q}}
\def\dddotq{q^{(3)}}

\def\ac{\ddotq}
\def\j{\dddotq}

\def\qdm1{\q_{k-1}}
\def\vdm1{\dotq_{k-1}}
\def\acdm1{\ac_{k-1}}
\def\jdm1{\j_{k-1}}

\graphicspath{{./img}}

\newcommand\black[1]{{\color{black} #1}}

\usepackage{xcolor} % per testo grigio/bordi

\makeatletter  % Edit the following values as appropriate
\def\@IEEEBIOphotowidth{0.8in}    % width of the biography photo area
\def\@IEEEBIOphotodepth{1in}   % depth (height) of the biography photo area
\def\@IEEEBIOhangwidth{0.85in}    % width cleared for the biography photo area
\def\@IEEEBIOhangdepth{1in}    % depth cleared for the biography photo area
\makeatother

\begin{document}
\title{Optimizing Design and Control Methods\\ for Using Collaborative Robots\\ in Upper-Limb Rehabilitation}
\author{Dario Onfiani, Marco Caramaschi, Luigi Biagiotti, Fabio Pini
\thanks{All the authors are with the Department of Engineering ``Enzo Ferrari'', University of Modena and Reggio Emilia, via Pietro Vivarelli 10, 41125 Modena, Italy, e-mail: \{dario.onfiani,marco.caramaschi,luigi.biagiotti,fabio.pini\}@unimore.it.}% <-this % stops a space
\thanks{\black{This reasearch involved human subjects. % in its research. ù
Approval for all ethical and experimental procedures and protocols was granted by the Ethics Commission of the University of Modena and Reggio Emilia under Application No. 310306.}}
\thanks{This work was supported in part by project PRIN 2020 Co-MiR, funded by MUR,  CUP E93C20007880001,  and by project Fit For Medical Robotics, funded by MUR, funding program PNRR-PNC, CUP B53C22006810001.}
% \thanks{The final version of this research has been published in \textit{IEEE/ASME Transactions on Mechatronics (DOI:\href{https://doi.org/10.1109/TMECH.2025.3567128}{10.1109/TMECH.2025.3567128})}. Readers are encouraged to cite this article when referring to this work:\\[0.5em]
% \texttt{@article\{onfiani2025optimizing,\\
% \ \ title=\{Optimizing Design and Control Methods for Using Collaborative Robots in Upper Limb Rehabilitation\},\\
% \ \ author=\{Onfiani, Dario and Caramaschi, Marco and Biagiotti, Luigi and Pini, Fabio\},\\
% \ \ journal=\{IEEE/ASME Transactions on Mechatronics\},\\
% \ \ year=\{2025\},\\
% \ \ publisher=\{IEEE\} \}}.}
}

\makeatletter
\begin{titlepage}
  \centering

  % --- Box unico con IEEE notice + DOI + BibTeX ---
  \setlength{\fboxsep}{10pt}%
  \setlength{\fboxrule}{0.6pt}%
  \noindent\fbox{%
    \begin{minipage}{0.96\linewidth}
      % --- IEEE Notice ---
      \textbf{IEEE Copyright Notice}\par
      \vspace{0.4em}
      The final version of record is available at DOI:
      \href{https://doi.org/10.1109/TMECH.2025.3567128}{\textbf{DOI:10.1109/TMECH.2025.3567128}}.\par
      Copyright © IEEE. Personal use is permitted. 
      For other uses, please contact pubs-permissions@ieee.org.\par
      \textcolor{gray!70}{This author-archived version is provided for open access.}

      \vspace{1.2em}

      % --- BibTeX citation ---
      \textbf{Bib\TeX{} citation:}
      \vspace{0.3em}

      %\setlength{\fboxsep}{6pt}%
      %\setlength{\fboxrule}{0.4pt}%
      %\fbox{%
        \begin{minipage}{0.92\linewidth}
          \ttfamily
          @article\{onfiani2025optimizing,\\
          \ \ title=\{Optimizing Design and Control Methods for Using Collaborative Robots in Upper Limb Rehabilitation\},\\
          \ \ author=\{Onfiani, Dario and Caramaschi, Marco and Biagiotti, Luigi and Pini, Fabio\},\\
          \ \ journal=\{IEEE/ASME Transactions on Mechatronics\},\\
          \ \ year=\{2025\},\\
          \ \ publisher=\{IEEE\}\\
          \}
        \end{minipage}
      %}
    \end{minipage}
  }

  % \vspace{2.0cm}

  % % --- Titolo ---
  % {\huge\bfseries \@title \par}

  % \vspace{1em}

  % % --- Autori ---
  % {\large \CoverAuthors \par}

  % \vspace{1.5em}

  % % --- Journal reference / DOI ---
  % \begin{minipage}{0.9\linewidth}
  %   \small
  %   \textbf{Journal reference:} \CoverJournalRef \\
  %   \textbf{DOI:} \href{https://doi.org/10.1109/TMECH.2025.3567128}{DOI:10.1109/TMECH.2025.3567128} \\
  %   \textbf{How to cite this work:} \CoverHowToCite
  % \end{minipage}

  \vfill

  % --- Nota finale ---
  \begin{minipage}{0.9\linewidth}
    \footnotesize
    This manuscript version may differ from the final published version (typesetting, minor edits).
    Please cite the journal version as the record of reference.
  \end{minipage}

\end{titlepage}
\makeatother
% ====== FINE COVER PAGE ======

\maketitle

\begin{abstract}
In this paper, we address the development of a robotic rehabilitation system for the upper limbs based on collaborative end-effector solutions. The use of commercial collaborative robots offers significant advantages for this task, as they are optimized from an engineering perspective and ensure safe physical interaction with humans. However, they also come with noticeable drawbacks, such as the limited range of sizes available on the market and the standard control modes, which are primarily oriented towards industrial or service applications. To address these limitations, we propose an optimization-based design method to fully exploit the capability of the cobot in performing rehabilitation tasks. Additionally, we introduce a novel control architecture based on an admittance-type Virtual Fixture method, which constrains the motion of the robot along a prescribed path. This approach allows for an intuitive definition of the task to be performed via Programming by Demonstration and enables the system to operate both passively and actively. In passive mode, the system supports the patient during task execution with additional force, while in active mode, it opposes the motion with a braking force. Experimental results demonstrate the effectiveness of the proposed method.%\\
\end{abstract}

\vspace{-5mm}
\section{Introduction}
In recent years, the use of robotic devices for post-operative rehabilitation has become increasingly common due to their numerous benefits. These devices are generally classified into two main categories: end-effector devices and exoskeletons \cite{gassert2018rehabilitation}.
The latter category is particularly advantageous because it enables rehabilitation exercises to be performed with an exceptionally high level of repeatability by precisely controlling each individual joint of the limb. Exoskeleton devices such as ARMIN \cite{nef2009armin}, Rupert \cite{sugar2007design}, and NESM \cite{crea2016novel} serve as excellent examples of this. However, these devices are hindered by their highly complex structure, which can be challenging to design effectively, and their limited adaptability to various exercise routines.
For these reasons, in this paper, we propose a solution based on collaborative robot manipulators (cobots), which falls under the end-effector category. End-effector solutions, such as MIT-MANUS \cite{krebs2013rehabilitation}, GENTLE/S \cite{loureiro2003upper}, REHAROB \cite{toth2009safe}, PUParm \cite{bertomeu2018human}, and EULRR \cite{zhang2020development}, offer several notable advantages.
 In occupational therapy contexts, the manipulators can guide the patient's limb along predetermined paths, replicating specific activities of daily living (ADL) \cite{mehrholz2012electromechanical,Maciejasz2014}. More generally, these systems allow for the free definition of the exercise path, providing multiple options for therapists to choose from while using the same device. Furthermore, these solutions offer the opportunity to adjust the level of effort required during rehabilitation, allowing adaptation of the exercise based on the patient's progress \cite{qian2015recent}.\\
The integration of cobots in such types of systems can significantly simplify their implementation, thus reducing the time required for clinical applications; however, it poses novel challenges. In particular, the design process must account for the kinematic architecture of commercial cobots,
which has been developed for general-purpose applications, and their limited payload. Additionally, to fully utilize the flexibility offered by these robotic solutions, a suitable control architecture is required to regulate the physical interaction between the robot and the human.\\  % Impedance control is one of the most commonly implemented solutions for guiding patients during task execution. This type of control sets the behavior of the robot to be compliant along the trajectory set by the therapist while remaining rigid in directions outside it \cite{ficuciello2014cartesian,keemink2018admittance,hogan1985impedance}. This approach offers the advantage of reducing the patient's spatial degrees of freedom to the path defined by the therapist within the robot's workspace. To simplify the robot's programming process, even for non-experts in robotics, the patient's trajectory can be defined using a Learning by Demonstration (LbD) approach \cite{lauretti2017learning,aleotti2006robust}. Under this paradigm, the therapist guides the robot's end-effector to record the exercise path that the patient will perform. Next, the recorded trajectory is encoded using tools like parametric functions (B-spline curves, Nurbs, etc. \cite{TRJBOOK}) or Dynamic Movement Primitives (DMPs) \cite{ijspeert2013dynamical} to obtain a compact and smooth representation of the user's motion.\\
\black{ In this paper, we present a novel end-effector rehabilitation system that utilizes a collaborative robot connected to the patient's limb via a simple mechanical interface. A key contribution of this work is the \textit{systematic integration of the robot} within a rehabilitation framework, significantly enhancing its capacity to support and guide patient movements.\\This contribution enables the development of a detailed, structured approach for configuring the rehabilitation scenario. Additionally, this approach facilitates the identification of the optimal configuration within the robot's workspace, allowing clinicians to position the robot maximizing its effectiveness in delivering the exercise. \\ Our primary contribution, however, is the \textit{design of an innovative control architecture} specifically tailored for rehabilitation robots. The control algorithm we developed enhances the capabilities of a general-purpose collaborative robot, transforming it into a versatile rehabilitation tool that can work in different modalities (standard, assistive, resistive) depending on the patient's needs and therapeutic goals.\\
%The paper is organized as follows. In Sec. \ref{sec:LiteratureOverview}, an analysis of the relevant literature is provided, and the features of a robotic rehabilitation system are identified. In Sec. \ref{sec:Optimization}, a procedure for optimizing the robot's pose is illustrated. Then, in Sec. \ref{sec:ControlArchitecture}, the novel control architecture of the robot for executing the task demonstrated by the therapist is described and experimentally validated in Sec. \ref{Sec.Experimental_results}. Finally, the conclusions of this research work are presented in Sec. \ref{sec:Conclusions}.
The paper is organized as follows. In Sec. \ref{sec:LiteratureOverview}, an analysis of the relevant literature is provided, and the features of a robotic rehabilitation system are identified. In Sec. \ref{sec:Optimization}, a procedure for optimizing the robot's pose is illustrated. Then, in Sec. \ref{sec:ControlArchitecture}, the novel control architecture of the robot for executing the task demonstrated by the therapist is described. In Sec. \ref{sec:ExperimentSetup}, the experimental setup and the task specification, which involve Learning by Demonstration (LbD), are detailed. The experimental validation of the proposed methodology is presented in Sec. \ref{Sec.Experimental_results}. Finally, the conclusions of this research work are presented in Sec. VII.% \ref{sec:Conclusions_real}.% \ref{sec:Conclusions}.
\vspace{-5mm}
\section{Related Works and Paper Contributions}
\label{sec:LiteratureOverview}
The design of the proposed control architecture has been guided by an in-depth analysis of the literature, aiming to identify the functional and technical features that contribute to an effective and flexible rehabilitation system.
From a functional perspective, rehabilitation tools are categorized according to the level of assistance they provide \cite{Maciejasz2014}, ranging from passive to active devices. This classification depends on whether the devices only offer resistive forces or can actively apply forces to the patient. Robotic devices enable the adjustment of the resistance experienced by the user during rehabilitation exercises and can incorporate an assist-as-needed (AAN) mechanism to help the patient complete the exercises \cite{zhang2020development,asNeeded2100}.
It is important to note that, in general, the terms ``active" and ``passive" refer to the behavior of the patient.
In the context of rehabilitative robots, passive therapy can be easily implemented because there is no need for feedback by the patient, whose limb is simply moved along predefined trajectories without the patient having to exert any effort. Accordingly, from a control perspective, this application involves simple trajectory tracking of a time-dependent trajectory, possibly with different stiffness gains along the free and constrained directions \cite{Tamantini2022}.
On the contrary, active therapy requires the robot end-effector to constrain the motion of the user's limb along specific directions while leaving movement free in other directions. In any case, it is the patient, possibly aided by the system, who must apply the force necessary to move the limb. In this scenario, the control must implement a so-called guiding \textit{Virtual Fixture} (VF), which does not depend on time but only on the geometric characteristics of the constraints \cite{Abbott2007,Bowyer2014}.
%With respect to this issue, as mentioned in \cite{saveriano2021dynamic}, the time dependency is a major limitation of DMPs, which are commonly used to encode the motions demonstrated by a user and easily adaptable to different situations/users. For this reason, in \cite{Tamantini2022}, a control architecture for orthopaedic robot-aided rehabilitation mixing DMPs with a geometric description of the constraints has been proposed.
Basically, guiding VFs can be implemented in two complementary ways:
\begin{enumerate}
\item The user interacts with the robot, which is subject to a force/velocity field that tends to maintain it on the desired path.
\item The user interacts with a virtual point, called {\it proxy}, that is constrained to move on the desired path, and then the robot, possibly connected through a virtual spring, tracks this point.
\end{enumerate}
Both approaches have their pros and cons.
Consider the scheme of \Fig{guidingVF}, where a generic guiding VF is defined by the parametric curve $\boldphi(s)$.
\begin{figure}[tb]
\vspace{-8mm}
\centering
\psfrag{x}[c][c][1][0]{ $x$}
\psfrag{s}[c][c][1][0]{ $s^\star$}
\psfrag{A}[c][c][0.8][0]{ $s=0 $}
\psfrag{B}[c][c][0.8][0]{ $s=l$}
\psfrag{f}[c][c][1][0]{ $\boldphi(s)$}
\psfrag{F}[c][c][1][0]{ $F_\parallel$} %width=0.3\columnwidth,height=2cm
\psfrag{H}[c][c][1][0]{ $F_{h}$}
    {\includegraphics[scale=0.55]{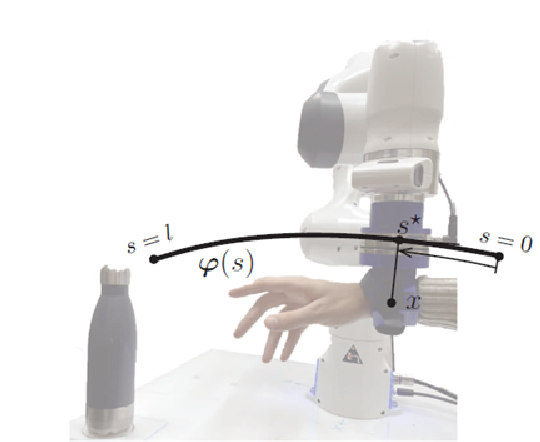}}
%%%%%%%%%%%%%%%%%%%%%%%%%%%%%%%%%%%
        %% OLD SECTION %%
%%%%%%%%%%%%%%%%%%%%%%%%%%%%%%%%%%
%\caption{Guiding virtual fixture on a generic curve $\boldphi(s)$.}
%%%%%%%%%%%%%%%%%%%%%%%%%%%%%%%%%%%
         %%END  OLD SECTION %%
%%%%%%%%%%%%%%%%%%%%%%%%%%%%%%%%%%%

\caption{\black{Guiding virtual fixture on a generic curve $\boldphi(s)$ parametrized by its arc-length, defining a rehabilitation task.}}

\label{fig:guidingVF}
\end{figure}
According to the former technique, described e.g. in
\cite{Pezzementi2007,Selvaggio2018,Papageorgiou2020},  it is necessary to compute at each time-stamp the value of parameter $s$, that minimize the distance between the curve $\boldphi(s)$ and the end-effector position $\x(t)$:
\begin{equation}
s^\star = \arg \min_{s\in[0,l]} \| \x(t)- \boldphi(s)\|.
\label{eq:s_computation}
\end{equation}
The solution of \eqref{eq:s_computation} based on iterative methods can be computationally expensive, thus limiting the maximum sampling rate of the digital control implementing the VF, with repercussions on the stability of the system itself \cite{Diolaiti2006}. Furthermore, the search for the nearest point on the path may yield more than one solution \cite{Selvaggio2018}.  For this reason, the approaches based on this type of solution are affected by \textit{singularities} in the motion representation, that may require the imposition of geometric constraints on the reference curve, such as its curvature \cite{Zhang2021}. The latter method is more efficient from a computational point of view and is not subject to singularity conditions due to the specific parameterization of the curve $\boldphi(s)$. However, it introduces into the system a new dynamics that may cause instability and unwanted effects, such as additional  elasticities.\\
 The problem of \textit{stability} for the overall system, composed of the robotic device physically interacting with the human, is crucial for guaranteeing the correct operation and safety of the application. Our work provides rigorous stability proofs for the control algorithms, a feature that is not always addressed in other works. For instance, \cite{zhang2015passivity} and \cite{asNeeded2100} do provide stability proofs, but several other systems, such as \cite{najafi2020using} and \cite{zhang2020development}, do not explicitly demonstrate stability. The lack of proof may pose risks in clinical applications, where patient safety is the priority.\\
 Another important feature of the system is the ability to easily define motions related to specific rehabilitation tasks. Obviously, the path $\varphi(s)$ can be expressed analytically. While this approach enables the precise generation of movement trajectories, it can be challenging for non-experts in robotics to implement and may not be well-suited for defining complex rehabilitation tasks.
For this reason, the patient's trajectory can be defined using a LbD approach \cite{lauretti2017learning,aleotti2006robust}. In this paradigm, the therapist guides the robot's end-effector to record the exercise path that the patient will perform. The recorded trajectory is then encoded using parametric functions (e.g., B-spline curves, NURBS, etc.) or Dynamic Movement Primitives (DMPs) \cite{ijspeert2013dynamical}, resulting in a compact and smooth representation of the user's motion.
% In contrast, LbD offers a simpler and more intuitive way for therapists to program the robot by demonstrating the desired movements. The combination of these two approaches enhances the system’s adaptability across a variety of rehabilitation tasks \cite{rossa2021robotic, zhang2015passivity}.
Reviewing existing systems, some, like  \cite{asNeeded2100, zhang2020development, Zhang2021} and \cite{najafi2020using} rely solely on analytic methods but lack the flexibility provided by LbD. Conversely, \cite{lauretti2017learning} employs the DMP architecture and relies exclusively on LbD but does not benefit from the precision of analytic methods.\\
A final important functional distinction among the various robotic rehabilitation systems proposed in the literature concerns the type of Virtual Fixture (VF) implemented. Broadly, these can be categorized as \textit{hard}  (or position-based) VFs, which ensure the patient closely follows the desired trajectory -making them valuable for tasks requiring high precision and repeatability, or \textit{soft} (or force-based) VFs, which allow the patient to deviate from the desired path, with the advantage of lower exchanged forces. Existing solutions, such as those presented in \cite{asNeeded2100, rossa2021robotic, Zhang2021, zhang2020development, najafi2020using, zhang2015passivity}, primarily focus on force constraints. These constraints are often based on potential or velocity fields, guiding the patient's motion along the correct path by restricting their movements. Additionally, \cite{lauretti2017learning} incorporates a form of position constraints designed to minimize errors introduced by the user during task execution.\\
\begin{table}[tb]
\centering
\large
\resizebox{\columnwidth}{!}{%
 \begin{tabularx}{\textwidth}{@{\extracolsep{\fill}}cccccccccc}
        %\hline
        \cline{1-9}
        \rule[-1.25ex]{0pt}{5ex}
        \textbf{Ref.} & \multicolumn{2}{c}{\textbf{\parbox{2cm}{\centering Interaction Modes}}} & \multicolumn{2}{c}{\textbf{\parbox{2cm}{\centering Curve Definition}}} & \multicolumn{2}{c}{\textbf{\parbox{2cm}{\centering Constraint Definition}}} & \multicolumn{1}{c}{\textbf{\parbox{1.5cm}{\centering Stability proved}}} & \multicolumn{1}{c}{\textbf{\parbox{2cm}{\centering Singularity Free}}} \\[3mm]  %\cline{1-9}
                  & Active & Passive & Analytic & LbD & Soft & Hard &  &  \\ \cline{1-9}
         \rule[-1.25ex]{0pt}{4ex} \cite{asNeeded2100}    &  \ding{51}     &    \ding{51}     &   \ding{51}       &    &   \ding{51}    &    & \ding{51}  &  \\ \cline{1-9} %21
         \rule[-1.25ex]{0pt}{4ex} \cite{rossa2021robotic}    &         &  \ding{51}        &   \ding{51}        &  \ding{51}    &  \ding{51}      &      &  \ding{51}   &   \\ \cline{1-9} %31
         \rule[-1.25ex]{0pt}{4ex}\cite{zhang2020development}    &   \ding{51}      &          &   \ding{51}        &      &  \ding{51}      &      &     &   \\ \cline{1-9} %10
         \rule[-1.25ex]{0pt}{4ex}\cite{Zhang2021}  &   \ding{51}     &    \ding{51}      &   \ding{51}        &      &  \ding{51}      &      &     &   \\ \cline{1-9} %30
          \rule[-1.25ex]{0pt}{4ex}\cite{lauretti2017learning}  &         &    \ding{51}      &   \ding{51}    &     &        &   \ding{51}   &      & \ding{51}  \\ \cline{1-9} %17
           \rule[-1.25ex]{0pt}{4ex}\cite{najafi2020using}  &       &    \ding{51}      &   \ding{51}    &     &  \ding{51}      &      &  \ding{51}   &   \\ \cline{1-9} %32
          \rule[-1.25ex]{0pt}{4ex} \cite{zhang2015passivity}  &    \ding{51}     &    \ding{51}      &   \ding{51}    &  \ding{51}   &  \ding{51}      &      &     &    \\ \cline{1-9} %33
                    \rule[-1.25ex]{0pt}{4ex} Our  &    \ding{51}     &    \ding{51}      &   \ding{51}    &  \ding{51}   &  \ding{51}      &   \ding{51}   & \ding{51}    & \ding{51}   \\ \cline{1-9}
    \end{tabularx}
    }
    \caption{\label{Tab:LiteratureReview} Main features of the control architectures for end-effector type rehabilitation robots.}
    \vspace{-5mm}
\end{table}
In Table \ref{Tab:LiteratureReview}, some notable research works focused on the control of end-effector type rehabilitation robots have been classified according to the key features identified above. Many of the proposed control architectures possess most of the desirable characteristics that such a system must have, but not all of them. The goal of the proposed research is to integrate all these features into a single framework.
The resulting solution is an \textit{Admittance-type Virtual Fixture control} that transforms the system into a passive mechanical tool, which can be easily programmed by therapists through a Learning by Demonstration approach. Additionally, our architecture facilitates \textit{adaptive force application}, enabling the robot to assist or resist the patient as needed during the exercise.
Interestingly, due to the particular expression of the elastic function embedded in the admittance model, it is possible to separately define the restoring force level and the maximum deviation from the reference curve, combining soft and hard constraints.

%Our system also demonstrates notable advancements in \textit{constraint definition}, offering both \textit{force-based} and \textit{position-based} constraints. This dual approach provides a versatile and adaptive control strategy, allowing the robot to follow a predefined path specifically tailored to the rehabilitation task. Position-based constraints are particularly valuable for tasks requiring high precision and repeatability, ensuring the patient closely follows the desired trajectory. In contrast, force-based constraints introduce "soft constraints," offering greater flexibility and tolerance during task execution.
}

\vspace{-4mm}
\section{Robot-trajectory relative pose optimization}
\label{sec:Optimization}
A methodology for designing cobot-assisted rehabilitation solutions is proposed. Based on the physiotherapist's recommendations, the layout of rehabilitation exercises can be designed by selecting the appropriate cobot and end-effector and optimizing the placement of motion trajectories within the workspace.Shifting the perspective, we aim to determine the robot pose that maximizes a given design criterion, starting with a specific type of exercise. In our previous work \cite{caramaschi2022workspace}, the optimization of the robot pose was based on the manipulability index:
\[
m(\q) = \sqrt{\det\!\big(\J(\q)\J^T(\q)\big)}
\]
where $\J(\q)$ denotes the Jacobian Matrix of the manipulator. Based on the index $m(\q)$, the cobot workspace is divided into regions that guarantee a minimum value, identifying those zones where the robot is capable of exerting forces in all Cartesian directions without causing excessive values of the joint torques. However, considering that the primary limitation for the cobots available on the market is represented by the payload they can sustain, and that in the early stages of the rehabilitation process, the patient can barely support the weight of their arm, thus resulting in the main component of the force acting on the cobot when they grasp the end effector being in the vertical direction, a different index has been preferred. Based on the above considerations, the new index evaluates the maximum net force the robot can exert along the $z$ axis of the task space\\
Let us consider the Euler-Lagrange model of a robot manipulator interacting with a human:
\begin{equation}
\M(\q)\ddot{\q} + \C(\q,\dot{\q})\dot{\q} +\boldsymbol{g}(\q) = \boldsymbol{\tau}+ \J^{T}(\q)\F_{h}
\label{eq:RobotDynamics}
\end{equation}
where $\q \in \mathbb{R}^n$ is the vector of the joint variables,  $\q(\q)$ is the inertia matrix, $\mathbf{C}(\q, \dot{\q})\dot{\q}$ is the vector of Coriolis/centrifugal torques,  $\mathbf{g}(\q)$ is the gravitational torques vector,  $\boldsymbol{\tau}$ is the actuator torque vector,  and
$\boldsymbol{\tau}_h = \J^T(\q)\F_h$ are the joint torques resulting from the external wrench $\F_h$ applied by the patient to the end-effector.\\
In static conditions, i.e., when $\ddot{\q}=\dot{\q}=0$, and considering that the force generated by humans is oriented along the $-z$ direction, $\F_h = [0,\,0,\,-\!F_z,\,0,\,0,\,0]^T$, the actuator torques required to statically balance the robot at the configuration $\q$ are given by
\[
\tau_i = \J_{3,i}(q) F_z+\g_i(q),\hspace{6mm} i=1,\ldots,n
\]
where $\J_{3,i}(q)$ is the element in position $(3,i)$ of the robot Jacobian, and $\g_i(q)$ is the $i$-th component of the gravitational torques vector. By imposing the actuation limits $\left| \tau_i \right| \leq \tau_{lim,i},, i = 1,...,n$, a bound on the maximum vertical force $F_z$ that the robot can resist is deduced as follows:
\be
F_{z,\max} =\min_{i} \left\{ \frac{\tau_{lim,i} - |\g_i(q)|}{|\J_{3,i}(q) |} \right\}.
\label{eq:MaxVerticalForce}
\ee
\begin{figure}[tb]
{\includegraphics[width=0.3\columnwidth]{ 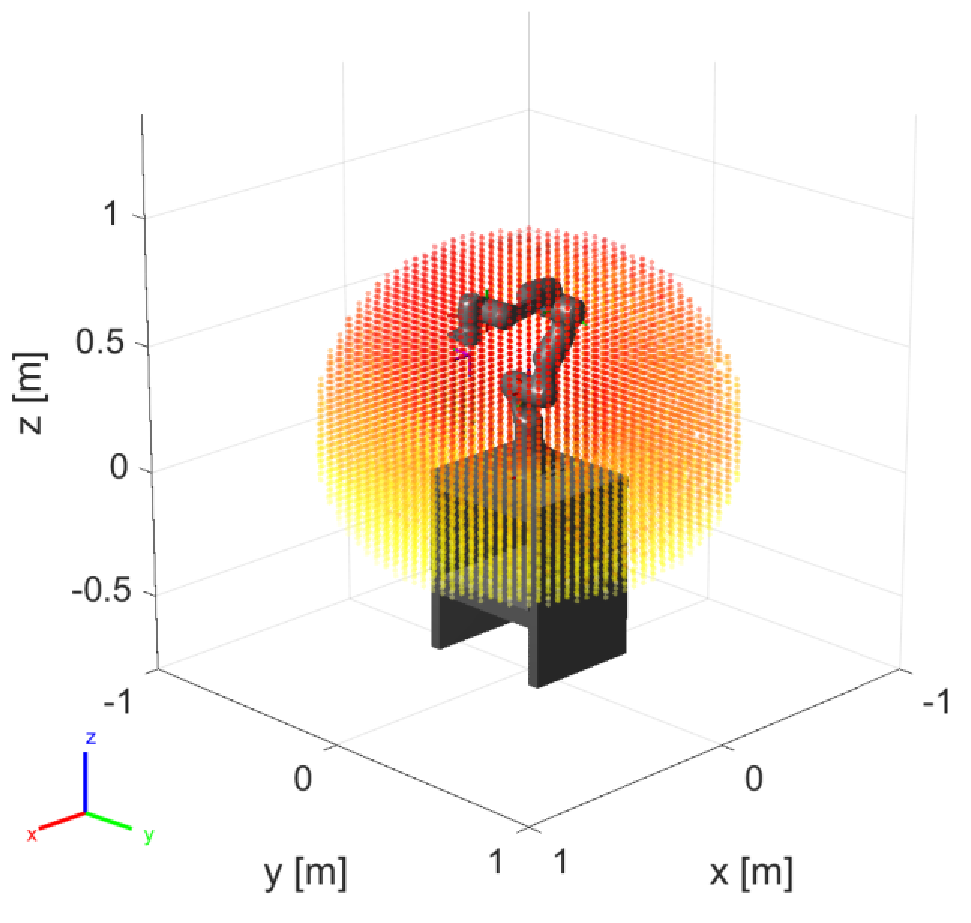}}
{\includegraphics[width=0.3\columnwidth]{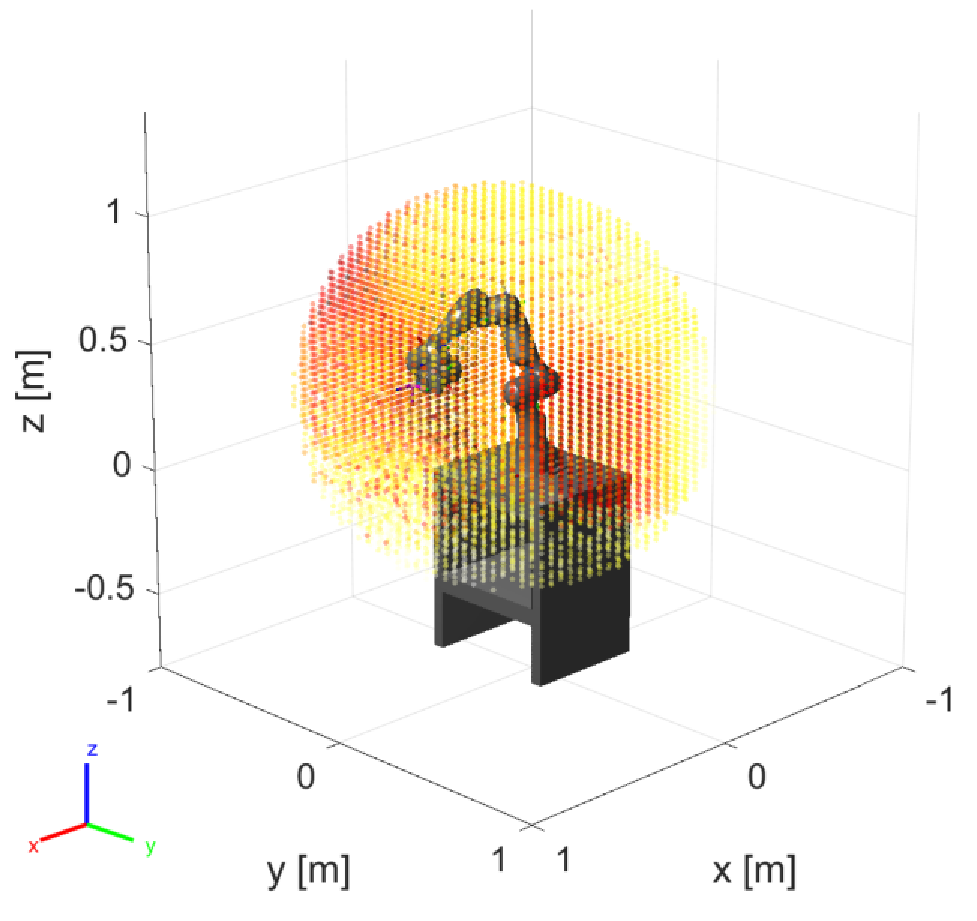}}
{\includegraphics[width=0.37\columnwidth]{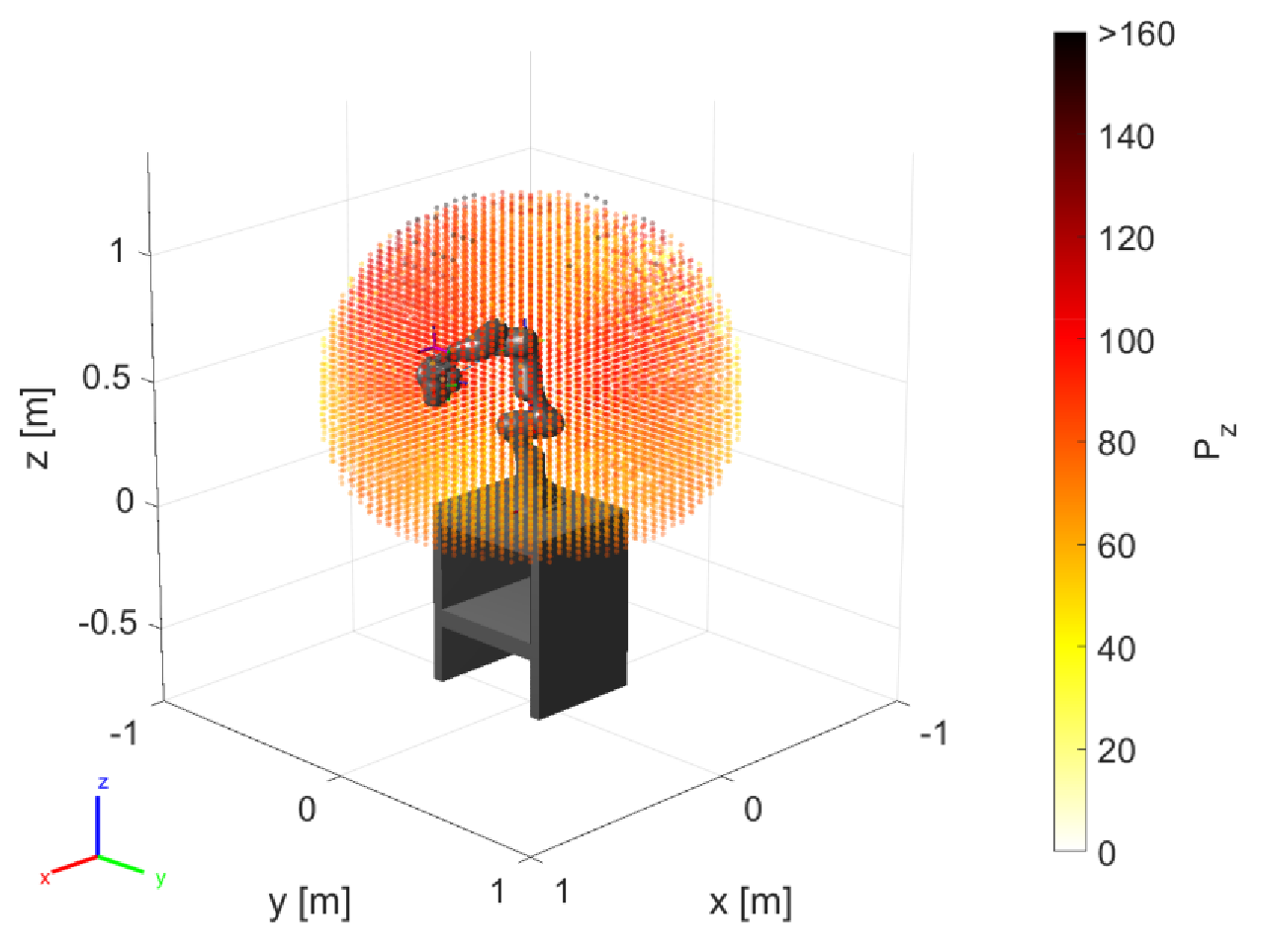}}\\
\small{(a)\hspace{2.5cm}(b)\hspace{2.1cm} (c)\hspace{3mm}}
	\caption{ Workspace maps based on the index $P_z(\x_i)$ with fixed orientations: Flange Down (a), Flange Horizontal (b) and Flange Up (c).}
 \label{fig:PointCloudRobot}
\vspace{-5mm}
\end{figure}
The force $F_{z,\max}$ is a scalar function of the robot configuration. By considering the inverse kinematic function $\q = \mbox{IK}(\p)$ of the manipulator, it can be used as an index for mapping the robot workspace. The pose of the manipulator $\p = [\x^T, \bgamma^T]^T$ is described by the position $\x$ and a minimal representation $\bgamma$ of the orientation of the robot end-effector. The 3D space is discretized with a finite number of points $\x_i$, and for a given constant orientation $\bar{\bgamma}$, each pose $\p_i$ is mapped into the corresponding robot configuration $\q_i$, and finally to the maximum force that the robot can apply at this point:
\[
P_z(\x_i):\,\,\,\p_i=\left[\!\begin{array}{c}
     \x_i\\
\bar{\bgamma}
\end{array}\! \right] \,\,\,\longrightarrow \,\,\, \q_i \,\,\, \longrightarrow \,\,\,{F_{z,\max}}_i.
\]
The function $P_z(\cdot)$ is called {\it payload index}.
A four-dimensional dataset $(\x_i, P_z(\x_i))$ is obtained. In Fig.\ref{fig:PointCloudRobot}, a color scale is associated with the payload values to create a 3D representation of the map for a given orientation of the flange. This procedure can be repeated for multiple flange orientations to obtain the desired maps. In the examples depicted in Fig.\ref{fig:PointCloudRobot}, three flange orientations have been considered: Flange Down (a), Flange Horizontal (b), and Flange Up (c). They illustrate the significant variation in payload capability resulting from the different orientations within various regions of the robot's workspace.

Workspace optimization involves determining the optimal placement of the desired trajectory relative to the cobot's base reference frame. For a given position $j$, the trajectory, discretized by considering an ordered set of points $\boldphi_k$, $k=1,\ldots, n$, is ranked based on the minimum value of the payload index, denoted as $\pi_j = \min_k {P_z(\boldphi_k)}$. By iteratively exploring various locations within the robot's workspace (the trajectory is translated in the $x$ and $y$ directions and rotated around its ``center of gravity'') and for different orientations of the end-effector, the optimal configuration is identified as
\[
\pi_{\mbox{\small opt}} = \max_j\{\pi_j\}.
\]
%The design methodology is summarized in Fig.\ref{fig:MethodDiagram}.
%{\includegraphics[width=0.8\columnwidth]{img/FigMeth2.eps}}
% \begin{figure}[tb]
% 	\centering
% 	{\includegraphics[width=0.8\columnwidth]{img/Method3.eps}}
% 	\caption{Design methodology for collaborative rehabilitation solutions.}
%     \label{fig:MethodDiagram}
% \end{figure}
% Following the steps in the diagram enables the design of an optimized layout for cobot-assisted rehabilitation solutions, starting from a specific exercise or a set of desired exercises.
Once the layout is defined, and the cobot is installed with the end-effectors in place, ensuring that the system is user-friendly for the therapist and easily customizable becomes crucial. To adjust exercises according to specific requirements or tailor movements to the patient's anthropometric dimensions, a LbD methodology has been devised. This involves the physiotherapist guiding the cobot's end-effector via kinesthetic interaction along the desired trajectory.

\vspace{-4mm}
\section{Control Architecture for Human-Robot Interaction}
\label{sec:ControlArchitecture}
A novel control architecture that mixes admittance control and guidance virtual fixtures is developed to constrain the motion of the cobot's end-effector along a 3D path specified by the therapist  without imposing a specific time law. In this way, it is the patient connected to the robot's end-effector who has to impose the movement along the curve by applying forces with the rehabilitated limb. \\
\begin{figure}[tb]
\centering
\psfrag{m}[c][c][0.9][0]{ $m,b$}
\psfrag{s}[c][c][0.9][0]{ $s^\star$}
\psfrag{A}[c][c][0.9][0]{ $s=0 $}
\psfrag{B}[c][c][0.9][0]{ $s=l$}
\psfrag{f}[c][c][0.9][0]{ $\boldphi(s)$}
\psfrag{F}[c][c][0.9][0]{ $\F_\parallel$} %width=0.3\columnwidth,height=2cm
\psfrag{H}[c][c][1][0]{ $\F_{h}$}
    % {\includegraphics[scale=0.5]{img/ConstrainedDynamics3.eps}}\\[-25mm]
    %
    {\includegraphics[width=0.7\columnwidth]{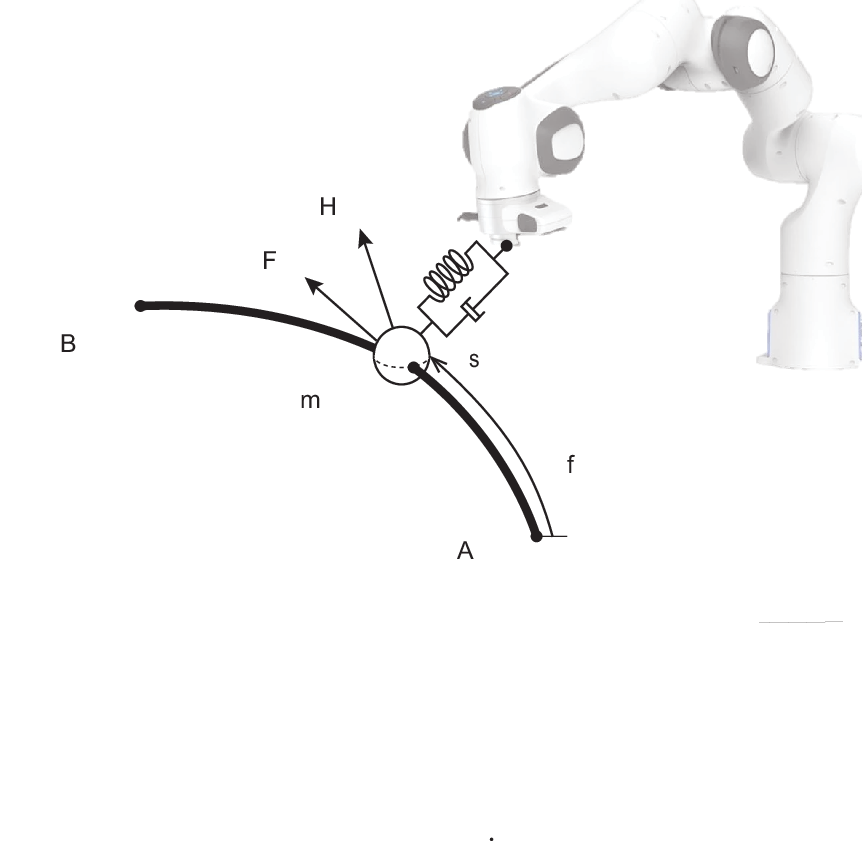}}\\[-15mm]
    % {\small (a)}\\[3mm]
    {\small (a)}\\[1mm]
\psfrag{F}[c][c][0.9][0]{ ${\boldsymbol{F_{h}}}$}
\psfrag{x}[c][c][0.9][0]{ ${\boldsymbol{\dot{x}}}$}
\psfrag{H}[c][c][0.9][0]{ Human}
\psfrag{R}[c][c][0.85][0]{ Robot}
\psfrag{P}[c][c][0.9][0]{ Power Port}
\psfrag{C}[c][c][0.85][0]{ Controlled robot}
% \psfrag{V}[c][c][0.8][0]{ {\centering Virtual Mass Dynamics}}
\psfrag{V}[c][c][0.7][0]{ {\centering Virtual Mass Dynamics}}
% \psfrag{I}[c][c][0.85][0]{ \parbox{2cm}{\centering Cartesian Impedance Control}}
\psfrag{I}[c][c][0.6][0]{ \parbox{2cm}{\centering Cartesian Impedance Control}}
% \psfrag{p}[c][c][0.77][0]{ \parbox{2cm}{\centering Path Generation $\varphi(s)$}}
\psfrag{p}[c][c][0.58][0]{ \parbox{2cm}{\centering Path Generation $\varphi(s)$}}
\psfrag{u}[c][c][0.9][0]{ $\boldsymbol{\hat{F}_{h}}$}
\psfrag{X}[c][c][0.9][0]{ $\x_{d}$, $\dot{\x}_{d}$, $\dot{\x}_{d}$}
\psfrag{e}[c][c][0.9][0]{ $\Dot{\x}_{d}$}
\psfrag{f}[c][c][0.9][0]{ $\ddot{\x}_{d}$}
\psfrag{T}[c][c][0.9][0]{ $\boldsymbol{\tau}$}
\psfrag{s}[c][c][0.55][0]{ $s^\star$}
\psfrag{S}[c][c][0.9][0]{ $s^\star$}
\psfrag{f}[c][c][0.55][0]{ $F_\parallel$}
\psfrag{m}[c][c][0.55][0]{ $m,b$}
% {\includegraphics[height=0.7\columnwidth,width=0.9\columnwidth]{img/BlockSchemeControl%4Rehab.eps}}\\[-4mm]
{\includegraphics[width=0.9\columnwidth]{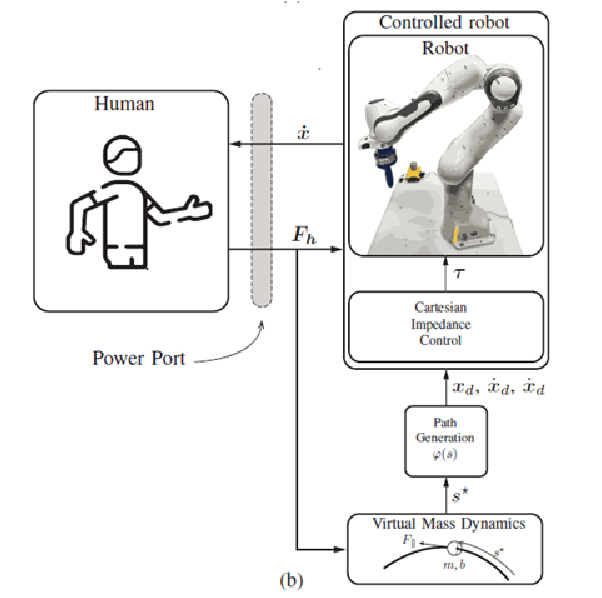}}\\[-4mm]
%{\small (b)}\\
\caption{Working principle of the proposed control architecture based on a constrained point-wise mass (a), and related block scheme representation (b).}
\label{fig:control_scheme}
\vspace{-5mm}
\end{figure}
As mentioned in the introduction, the basic idea, which relies on a virtual proxy \cite{Zilles1995,Abbott2007} affected by the force exchanged with the user, is quite common in haptic and human-robot interaction applications. However, the implementation we propose is novel and presents several advantages, which will be discussed in the following.\\
Let us assume that the reference path, denoted as $\boldsymbol{x_{d}}$, has been settled by the therapist using   a LbD procedure, and is defined by a parametric function:
\begin{equation}
\boldsymbol{x_{d}}= \boldphi(u).
\label{eq:CurveDefinition}
\end{equation}
 Function $\boldphi(\cdot)$ represents any curve with a proper degree of continuity, i.e. at least of class $\mathcal{C}^1$. In the experiments, we consider cubic B-spline functions, obtained  by interpolating the points registered when the user executes the robotic rehabilitation task according to a LbD approach.\\
In kinesthetic teaching, the independent variable $u$ is typically time (or some related function), ensuring that the trajectory $\x_d$ precisely replicates the demonstrated motion. However, since the proposed application aims to enforce a prescribed geometric path without specifying how the path is tracked, an arc-length parameterization should be considered. This can be achieved by composing the function $\boldphi(u)$ in (\ref{eq:CurveDefinition}) with the function $u = u(s)$, which defines how the independent variable $u$ changes with arc length $s$.
The function $u(s)$ can be obtained by inverting the function
\begin{equation}
s(u) = \int_{u_0}^u \left\|\frac{d\boldsymbol{\varphi}(\tau)}{d\tau}\right\|d\tau
\nonumber
\end{equation}
which can be easily computed numerically during the interpolation procedure of the data points recorded from the demonstrated trajectory \cite{GASPAR2018,Onfiani2022}. With a slight abuse of notation, we denote $\boldphi(s) = \boldphi(u(s))$.\\
As shown in \Fig{control_scheme}(a), a point-wise mass $m$ is constrained to follow the  $\boldphi(s)$ and moves subject to the force applied by the user to the robot end-effector. Simultaneously, the cobot needs to accurately track the position of the mass.
In \Fig{control_scheme}(b), a block-scheme representation of the overall control architecture is depicted. The Admittance Guiding Virtual Fixture is obtained by computing the forward dynamics of the virtual mass, considering the measured force $\hat \F_h$. This approach allows us to obtain the instantaneous value of the reference position, given by\\[-1mm]
\be\ x_d(t) = \boldphi(s(t)).
\label{eq:RefPosition}
\ee
 Note that, given $x_d(t)$, its derivative can be easily computed analytically as follows:
\beann
    \dot{\x}_{d}(t)&=&\boldphi'(s)\dot{s}(t) \\
    \ddot{\x}_{d}(t)&=&\boldphi''(s)\dot{s}^{2}(t)+\boldphi'(s)\ddot{s}(t)
\eeann
where  $ \boldphi'(s)=\frac{d\boldphi(s)}{ds} $ and
$ \boldphi''(s)=\frac{d^{2}\boldphi(s)}{ds^{2}}$.
In the following the dynamic behavior of the proxy and the position control of the robot are detailed.\\ \vspace{-9mm}%[-6mm]
\subsection{Dynamic equation of the virtual mass}
%{\it Dynamic equation of the virtual mass}\\
The dynamic model of the point mass $m$ is deduced by applying the standard procedure based on Lagrange equations. With the basic assumption that  the mass constrained to the curve $\boldphi(s)$ is not affected by gravity, the Lagrangian function equals the kinetic energy
\be \mathcal{L}=\mathcal{K} = \frac{1}{2}m \dot{\x}_{d}^T \dot{\x}_{d}= \frac{1}{2}m \dot s^2,\label{eq:KinEnergy}\ee
where the fact that, for the arc-length parameterization, $\|\boldphi'(s)\|=1$ has been exploited.
The Lagrange equation  for (\ref{eq:KinEnergy}) with respect to $s$ yields
\begin{equation}
m \ddot s +b \dot s  = F_\parallel
\label{eq:ConstrainedMassDynamics}
\end{equation}
where the $b \dot s$ is a non-conservative term taking into account the friction and $F_\parallel$ is the component of  force $\boldsymbol{\hat F}_h$ applied by the user to the robot tool (and detected by a force sensor) which is tangent to the curve, i.e.
\begin{equation}
%{F_\parallel} = \left.\textbf{T}\right|_{s=s^\star}\cdot \boldsymbol{F_{h}}
F_\parallel =  \boldphi'(s)^{T}\cdot  \boldsymbol{\hat F}_{h}
\label{eq:TangentialForce}
\end{equation}
where, because of the use of  arc-length parameterization, $\boldphi'(s)$ represents the unit tangent vector  to $\boldphi(s)$, at a generic point $s$.\\
The other components of the force $\boldsymbol{\hat {F}}_{h}$, that do not contribute to accelerating the mass, are compensated by the constraints.\\ Note that if $m\approx 0$ a proportional relationship between the tangent force and the velocity along the curve, which is the most common solution for Admittance-type VF \cite{Abbott2007}, is obtained. \\[0.5mm]
\vspace{-9mm}%\vspace{-7mm}
\black{\subsection{Cartesian impedance control of the robot}
Consider the dynamic model of a robot manipulator interacting with a human in  \eqref{eq:RobotDynamics}.
The control law
\be
\boldsymbol{\tau} = \M(\q) \y + \C(\q, \dot{\q})\dot{\q} +\boldsymbol{g}(\q)
\label{eq:ControlTorque}
\ee
with the auxiliary input
\be
\y=\J^{-1}(\q)\left( \ddot \x_d -\boldsymbol{\dot{J}}(\q)\dot \q +\K_D'\dot{\tilde{\x}} +\F_{el}'(\tilde{\x})\right)
\label{eq:AuxInput}
\ee
leads to the closed loop dynamics:
\begin{equation}
\ddot{\tilde{\x}} + \K_D'\dot{\mathbf{\tilde{\x}}} +\F_{el}'(\tilde{\x}) =\M_A^{-1}(\q)\F_h
\label{eq:RobotDynamics_controlled}
\end{equation}
where $\tilde{\x} = \x-\x_d$ represents the Cartesian error with respect to the desired pose $\x_d$. Finally, assuming:
\bea
\K_D'&=& \M(\q)_A^{-1}\K_D\\
%\mathbf{C_{A}'}(\q, \dot{\q}) &=& \M(\q)_A^{-1}\mathbf{C_{A}}(\q, \dot{\q})\\
\F_{el}'(\tilde{\x})  &=& \M(\q)_A^{-1}\F_{el}(\tilde{\x})
\eea
the impedance model of the robot in the Cartesian space becomes:\\[-2mm]
\begin{equation}
    \M_A(\q)(\q)\ddot{\tilde{\x}} + \K_D\dot{\tilde{\x}} + \F_{el}(\tilde{\x})
    = \F_h
    \label{eq:final_end_effector_dynamics}
\end{equation}
The matrix
 $\M_A(\q) = \J_A^{-T}(\q)\M(\q)(\q)\J_A^{-1}(\q)$ is the robot inertia in the Cartesian space, $\K_D>0$ is the desired damping matrix and, finally, $\F_{el}(\tilde{\x})$ is a generic elastic force, obtained by differentiating a scalar (potential) function  $U_{el}(\tilde{\x})\ge 0$, where $U_{el}(\tilde{\x})= 0$ if and only if $\tilde{\x}= 0$:\\[-2mm]
 \be \F_{el}(\tilde{\x})= \left(\frac{\partial U_{el}(\tilde{\x}) }{\partial \tilde{\x}}\right)^T.
 \label{eq:ElasticPotential}
 \ee
As highlighted in  \cite{Schaffer2003}, (\ref{eq:final_end_effector_dynamics}) represents a passive mapping from the external force $\F_h$ to the velocity error $\dot{\tilde{\x}}$, ensuring the stability of the system in feedback interconnection with a passive environment.\\
Finally, note that the inclusion of the acceleration $\ddot \x_d(t)$  in the control signal (\ref{eq:AuxInput}) justifies the adoption of second-order dynamics in (\ref{eq:ConstrainedMassDynamics}) to model the evolution of $s(t)$, that otherwise could lead to infinite values for this variable.}
\vspace{-5mm}% \vspace{-1mm}
\black{
\subsection{Stability Analysis}
\label{ssec:PassivityAnalysis}
 As illustrated in \Fig{control_scheme}(b), the overall control scheme is composed of three subsystems, namely Patient, Admittance Guidance VF and Controlled Robot. Its stability relies on the basic assumption that humans behave in a passive manner, at least when the position of their  limbs is kept constant \cite{Dyck2013,Anderson2023,haykin1970active}. We assume accordingly that  the human operator defines a passive velocity ($\dot \x$) to force ($-\F_h$) map.
% Therefore, to assure the stability of the overall system that forms a negative feedback interconnection with the user, it is sufficient to prove that the map from the force applied to the robot $\F_h$ and its velocity $\dot \x$ is passive as well. %This can be done, by considering for the dynamical system  composed by (\ref{eq:ConstrainedMassDynamics})  and (\ref{eq:final_end_effector_dynamics}), and characterized by the state vector $\xi=[\dot s,\,\tilde \x, \dot{\tilde \x} ]^T$, %$\xi=[s,\,\dot s,\,\tilde \x, \dot{\tilde \x} ]^T$, the storage function
% \begin{equation}
% S_r(\xi) = \frac{1}{2}m\dot{s}^{2} + \frac{1}{2}\boldsymbol{\dot{\Tilde{x}}}^T \M_{A}(\boldsymbol{q}) \boldsymbol{\dot{\Tilde{x}}} + U_{el}(\tilde{\x}).
% \label{eq:storage function_new}
% \end{equation}
To analyze the dynamical system described by (\ref{eq:ConstrainedMassDynamics}) and (\ref{eq:final_end_effector_dynamics}), with the state vector $\xi = [\dot{s}, \, \tilde{\boldsymbol{x}}, \, \dot{\tilde{\boldsymbol{x}}}]^T$,
% \[
% \xi = [\dot{s}, \, \tilde{\boldsymbol{x}}, \, \dot{\tilde{\boldsymbol{x}}}]^T,
% \]
we define the storage function as:
\begin{equation}
S_r(\xi) = \frac{1}{2}m\dot{s}^2 + \frac{1}{2}\dot{\tilde{\boldsymbol{x}}}^T \M_A(\boldsymbol{q}) \dot{\tilde{\boldsymbol{x}}} + U_{el}(\tilde{\boldsymbol{x}}).
\label{eq:storage function_new}
\end{equation}

The selected storage function serves as a valid Lyapunov function, allowing us to apply the Lyapunov theorem to assess the system's stability.\\
By noting that $\boldsymbol{\dot{\x}}_{d} = \boldphi'(s)\dot{s} $, $\Dot{\tilde{\x}} = \Dot{\x} - \Dot{\x}_{d}$, and assuming that the measured force matches the actual force (i.e., $\hat \F_h = \F_h$), we derive:
\begin{equation}
\dot{S}_r(\xi) = -b\dot{s}^2 - \Dot{{\tilde{\x}}}^{T} \K_{D} \Dot{\Tilde{\x}} + \Dot{{\x}}^{T} \boldsymbol{F}_{h}.
\label{eq:der_storage function3}
\end{equation}
Given that \( b > 0 \) and \( \K_{D} > 0 \),
\[\dot{S}_r(\xi) \le \Dot{{\x}}^{T} \boldsymbol{F}_{h}.\]
Thus, the system is passive with respect to the pair \( \langle \dot \x, \, \F_h \rangle \).

% \begin{figure}[h!]
%     \centering
%     \includegraphics[width=0.75\linewidth]{img/img_rev/scheme_stability_new2.eps}
%     \caption{Block scheme representation of the system on which we perform the stability analysis.}
%     \label{fig:Stability_scheme}
% \end{figure}
}
%%%%%%%%%%%%%%%%%%%%%%%%%%%%%%%%%%%
         %%END OLD SECTION %%
%%%%%%%%%%%%%%%%%%%%%%%%%%%%%%%%%%%

\vspace{-5mm}
\subsection{Quasi-static behavior}
Once it has been proven that the robotic system connected to the user is stable, it becomes of interest to consider the achievable performance in terms of position error and exchanged forces while the human is interacting with it. %By considering the characteristics of a typical rehabilitation task, that involves very small velocities and accelerations
Assuming that velocities and accelerations involved in a typical rehabilitation task are very small, we want to analyse the behaviour of the controlled robot, when $\dot \x \approx 0$ and $\ddot \x \approx 0$, that is close to an equilibrium state.
%Assuming
Considering that the user is exerting a constant force $\bar{\F}_h$, equations (\ref{eq:ConstrainedMassDynamics}) and (\ref{eq:final_end_effector_dynamics}) imply that at equilibrium\\[-4mm]
\bea
F_\parallel =  \boldphi'(\bar s)^{T}\cdot  \bar{\boldsymbol{F}}_{h} &=& 0 \label{eq:Orthogonality}\\
\F_{el}(\tilde{\x})  &=& \bar{\boldsymbol{F}}_h. \label{eq:ElasticForceBalance}
\eea
 Equation (\ref{eq:Orthogonality}) shows that the force applied by the user at the equilibrium must be orthogonal to the tangent vector to the desired curve $\bar{\x}_d = \boldphi(\bar{s})$, while equation (\ref{eq:ElasticForceBalance}) indicates that it must be counteracted by the elastic force acting on the robot, which is caused by the deviation $\bar{\x}$ from $\bar{\x}_d$.\\
  From an intuitive user perspective, the system's displacement, denoted by $\tilde \x$, should ideally align with the force causing it, similar to a standard mechanical spring. To achieve this, we can impose a specific structure on the elastic force function, $\F_{el}(\tilde{\x})$, as follows:
\begin{equation}
    \F_{el}(\Tilde{\x})=f_{el} (\|\Tilde{\x}\|)\,\frac{\tilde \x}{\|\tilde{\x}\|}
    \label{eq:generic_stiffness_function}
\end{equation}
where $f_{el} (\cdot)\ge0$ is a scalar function, such that $f_{el} (0)=0$.
In this way, the elastic force is always directed along $\tilde{\x}$ and its intensity is determined by $f_{el} (\cdot)$. However, the elastic connection plays different roles along the tangent and orthogonal directions to the curve. It needs to ensure minimal tracking error along the curve while still allowing patients to deviate from the planned path without encountering excessive forces. To achieve this, the elastic force has been decomposed into two complementary components:
\be \F_{el}(\Tilde{\x}) = \F_{el, \parallel}(\Tilde{\x}_\parallel)+\F_{el, \bot}(\Tilde{\x}_\bot)
\label{eq:FelDecomposition}\ee
where \\[-6mm]
\beann
\Tilde{\x}_\parallel &=& \boldphi'(\bar s)\boldphi'(\bar s)^{T}\,\Tilde{\x}\\
\Tilde{\x}_\bot &=&\left( \I_3-\boldphi'(\bar s)\boldphi'(\bar s)^{T}\right)\,\Tilde{\x}
\eeann
represent the tangent and orthogonal displacements to the curve at point $\boldphi(\bar{s})$, respectively.
Functions $\F_{el, \parallel}(\cdot)$ and $\F_{el, \bot}(\cdot)$ maintan the structure described in     \eqref{eq:generic_stiffness_function}. From \eqref{eq:FelDecomposition} and \eqref{eq:generic_stiffness_function}, the potential function that defines $\F_{el}(\tilde{\x})$ according to \eqref{eq:ElasticPotential} can be easily derived as:
\[U_{el}(\Tilde{\x}) = u_{el,\parallel}(\|\Tilde{\x}_\parallel\|)+u_{el,\bot}(\|\Tilde{\x}_\bot\|) \]
where  $\ds u_{el,\ast}(z) = \int \,f_{el,\ast} (z)dz$ is the primitive of $f_{el,\ast} (\cdot)$ such that $u_{el,\ast}(0)=0$.\\
If, for instance, it is assumed $f_{el,\ast}(z) = \kappa\,z$ where $\kappa$ is a positive constant, the constitutive equation of a standard linear spring is obtained, i.e.
 \be
\F_{el,\ast}(\Tilde{\x}_\ast)= \kappa\, \Tilde{\x}_\ast.
 \label{eq:LinearElasticity}
 \ee
This expression is used along the tangent direction, where a large value of $\kappa$ helps ensure minimal tracking error.  However, due to equations (\ref{eq:Orthogonality}) and (\ref{eq:ElasticForceBalance}), this component is not perceived by the user when moving slowly along the curve.

\begin{figure}[tb]
    \centering
    %\psfrag{F}[c][c][1][0]{$f_{el}(z)$ {\small [N]}}
    \psfrag{F}[c][c][1][0]{$f_{el}(\|\boldsymbol{\tilde{x}_{\bot}}\|)$ {\small [N]}}
    %\psfrag{x}[c][c][1][0]{$z$ {\small [mm]}}
    \psfrag{x}[c][c][1][0]{$\|\boldsymbol{\tilde{x}_\bot}\|$ {\small [mm]}}
    \psfrag{aaaaaaaaa}[l][l][0.6][0]{$\chi = 1$ \small{[N/mm]}}
    \psfrag{bbbbbb}[l][l][0.6][0]{$\chi = 2$ \small{[N/mm]}}
    \psfrag{cccccc}[l][l][0.6][0]{$\chi = 4$ \small{[N/mm]}}
    \includegraphics[width=0.8\columnwidth]{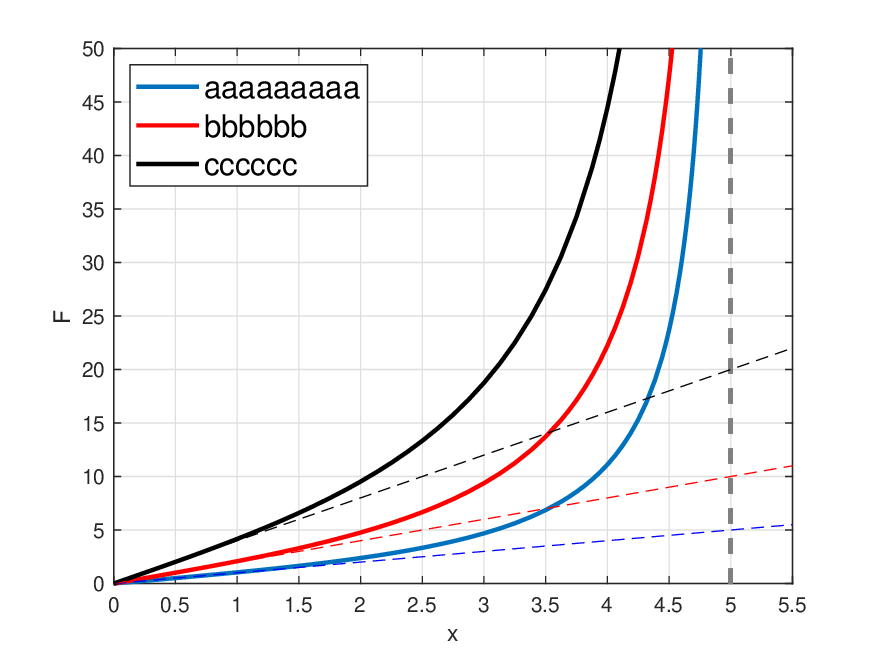}
    %\caption{ Nonlinear elastic function $f_{el}(z) = \chi \frac{ \delta^2 z}{\delta^2-z^2}$ for different values of parameter $\chi$ ($\delta = 5$ {\small [mm]}).
    \caption{Nonlinear elastic function $f_{el}(\|\boldsymbol{\tilde{x}_\bot}\|) = \chi \frac{ \delta^2 \|\boldsymbol{\tilde{x}_\bot}\|}{\delta^2-\|\boldsymbol{\tilde{x}_\bot}\|^2}$ for different values of parameter $\chi$ ($\delta = 5$ {\small [mm]}).}
    \label{fig:f_x_tilde}
    \vspace{-5mm}
\end{figure}

 % It is worth noting that (\ref{eq:LinearElasticity}) is a particular case of $\F_{el}(\Tilde{\x})= \K_p\, \Tilde{\x}$ with $\K_p = \mbox{diag}\{ \chi_i\}$ and the robot control (\ref{eq:final_end_effector_dynamics}) becomes a standard  Cartesian impedance control, where the natural inertia of the robot as felt in the workspace is maintained.\\
A more effective way to define a fixture for rehabilitation applications in the normal plane to the desired path is based on the function:
\be f_{el,\ast}(z) = \ds \chi \frac{ \delta^2 z}{\delta^2-z^2} \,\,\,\Rightarrow\,\,\, \F_{el,\ast}(\Tilde{\x}_\ast) = \ds \chi \frac{ \delta^2 }{\delta^2-\| \tilde{\x}_\ast\|^2} \tilde{\x}_\ast
 \label{eq:NonLinearStiffness}\ee
where  $\chi$ and $\delta$ are free parameters that respectively define the stiffness for small deformations ($z \approx 0$) and the maximum allowable displacement, as shown in \Fig{f_x_tilde}. When $z$ approaches $\delta$, the magnitude of the elastic force tends to infinity. In this way, the motion of the patient during the exercise is restricted to a maximum distance $\delta$ in the normal direction to the desired geometric path.\\
Interestingly, the potential function derived from (\ref{eq:NonLinearStiffness}), i.e.
% \be
% u_{el}(z) =  \frac{\chi \delta^{2}}{2}\log\biggl(\frac{\delta^{2}}{\delta^{2}-z^{2}}\biggr) \,\,\,\Rightarrow\,\,\, U_{el}(\Tilde{\x}) =  \frac{\chi \delta^{2}}{2}\log\biggl(\frac{\delta^{2}}{\delta^{2}-\| \tilde{\x}\|^{2}}\biggr)
%  \label{eq:NonLinearPotential}\ee
 \be
 U_{el,\ast}(\Tilde{\x}_\ast) =  \frac{\chi \delta^{2}}{2}\log\biggl(\frac{\delta^{2}}{\delta^{2}-\| \tilde{\x}_\ast\|^{2}}\biggr),
 \label{eq:NonLinearPotential}\ee
has the same form as standard Barrier Lyapunov Functions \cite{Tee2009}, used for preventing constraint violation in dynamic systems, as they grow to infinity when their argument approaches the given limit $\delta$.
\begin{figure}[tb]
\centering
\psfrag{f}[c][c][1][0]{ $\black{\F_{h}}$}
\psfrag{F}[c][c][1][0]{ $\black{\F_{\bot}}$}
\psfrag{P}[c][c][1][0]{ $\boldphi(s)$}
\psfrag{S}[c][c][1.25][0]{ $\bar{s}$}
%\psfrag{T}[c][c][1][0]{ $\black{\boldphi'(\bar s)}$}
\psfrag{T}[c][c][1][0]{ $\black{F_\parallel}$}
\psfrag{p}[c][c][0.8][0]{ $\F_{el}(\tilde{\x})$}
\psfrag{d}[c][c][1][0.8]{ $\K_{D}$}
%width=0.3\columnwidth,height=2cm
    {\includegraphics[width=0.7\columnwidth]{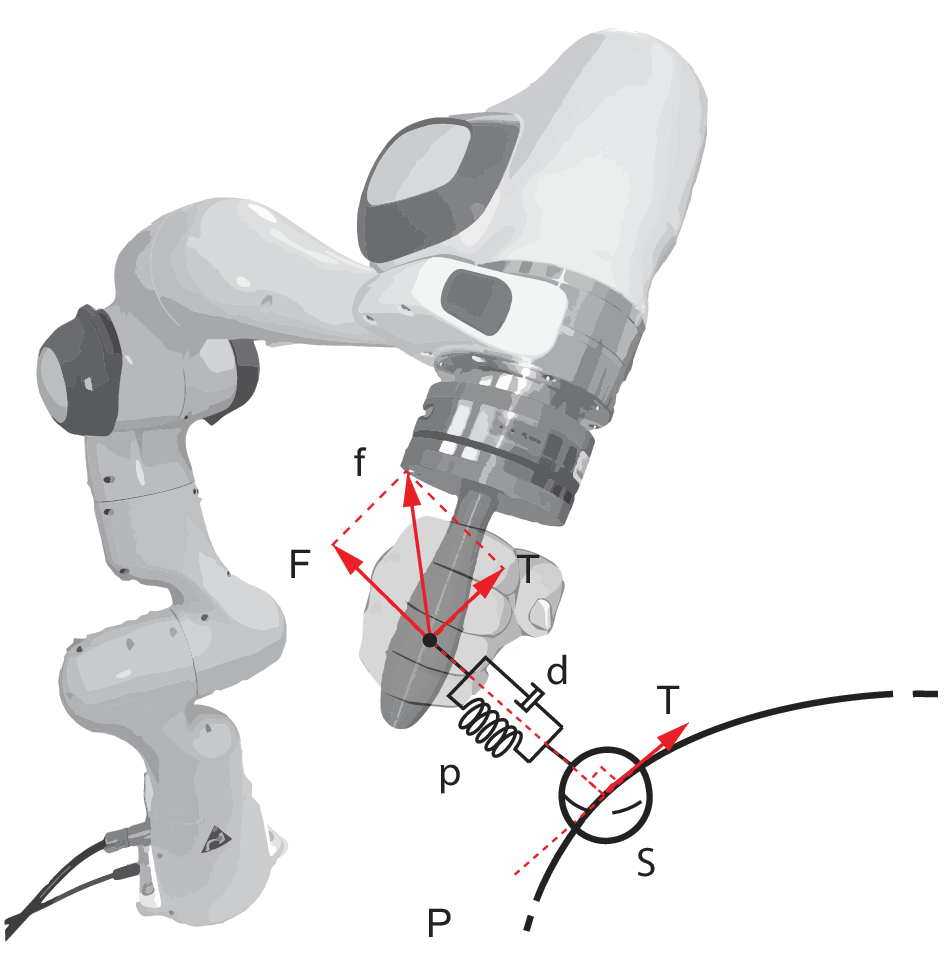}}
    %\vspace{-25mm}
\caption{ Equivalent mechanical system of the proposed controller in an equilibrium configuration.}
\label{fig:Equilibrium}
\vspace{-5mm}
\end{figure}
In Fig.\ref{fig:Equilibrium}, the behavior of the overall system is illustrated by means of an equivalent mechanical representation: while the human force $\F_h$ affects the robot, only its tangent component moves the mass along the desired path. At equilibrium, when this component is equal to zero, the normal component $F_\bot$ causes a deviation from the desired geometric path along the normal direction.
The control scheme obtained with the elastic function defined by
\[
\F_{el}(\Tilde{\x}) = \kappa\,\Tilde{\x}_\parallel \,+\,  \chi \frac{ \delta^2 }{\delta^2-\| \tilde{\x}_\bot\|^2} \tilde{\x}_\bot
\label{eq:FelTotal}
\]
is a specific instance of a band-type controller. This type of controller defines the boundaries of a virtual channel, where the motion of the human limb is constrained by forces applied in the normal direction \cite{zhang2020development,lin2020spatial}. In the proposed architecture, within the channel, there exists a residual elastic force towards the reference trajectory whose intensity can be freely chosen by adjusting $\chi$.\\
In this way, the movement of the human limb is not restricted to a specific trajectory \cite{zhang2015passivity} or a velocity profile \cite{asl2018assist, najafi2020using}, but is determined by the interaction between the user and the robot/virtual mass. The absence of a time constraint in the execution of the exercise makes the duration of the motion a useful parameter for estimating the functional ability of the subjects.

Moreover, the proposed scheme, based on Admittance Guiding Virtual Fixture, offers additional advantages, namely:
 \begin{itemize}
     \item The implementation, based on force-to-position causality, does not require knowledge of the normal direction to the desired reference path \cite{Ren2022}, whose estimation can be computationally expensive and, in some cases, even impossible since multiple solutions may exist, e.g., when a path has an intersection.
     \item Help-as-needed mechanisms or, conversely, resisting forces can be easily integrated into the control scheme by adding an appropriate virtual force in Equation (\ref{eq:ConstrainedMassDynamics}) governing the mass dynamics.
 \end{itemize}
\vspace{-2.5mm}
\section{Experimental Setup and Task Specification Using LbD}
\label{sec:ExperimentSetup}
The proposed methodology, which involves selecting the optimal robot configuration, programming the rehabilitative task through demonstration, and executing it via human-robot interaction, has been tested using a Franka Emika Panda, a collaborative robot with 7 degrees of freedom equipped with an Axia80-M20 force-torque sensor mounted on its terminal flange. \black{Since the robot is redundant, a torque vector defined in the null space of the Jacobian transpose, which maximizes the distance from the joint limits, has been added to the torque control defined by (\ref{eq:ControlTorque}), see \cite{SicilianoBook}.  %It is important to note that the robot is redundant, and as a result, it is necessary to consider its null space dynamics. The resolution of the redundancy problem stemming from the robot's structure is elaborated in Sec. \ref{redundancy}.
\begin{figure}[tb]
    \centering
        \centering
        \includegraphics[height=3.5cm]{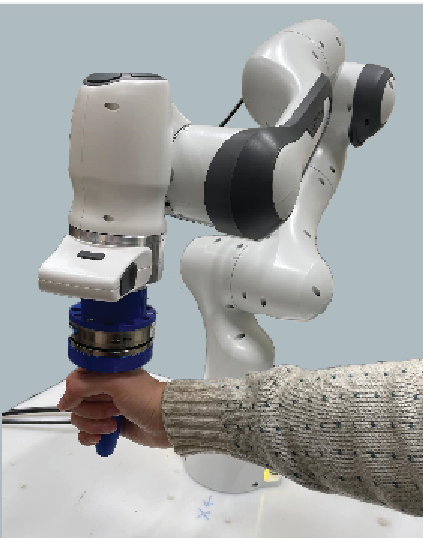}\hspace{2mm}
        \includegraphics[height=3.5cm]{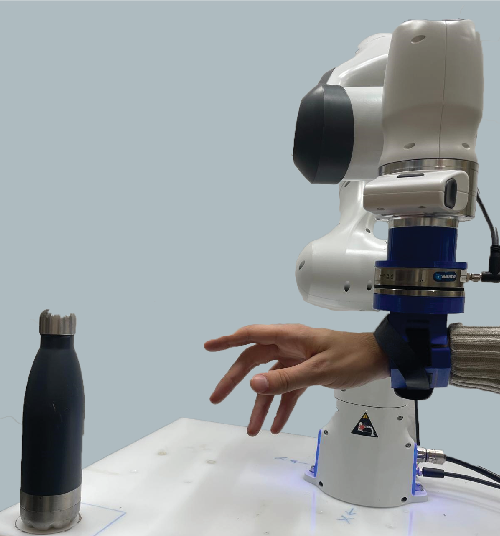}\\
        {\small (a) \hspace{3cm} (b)\hspace{2mm}}
        \caption{Experimental setup for basic experiments on human-robot co-manipulation tasks (a) and for the assisted execution of Activities of Daily Living (b).}
        \label{fig:ExperimentalSetup1}
        \vspace{-2mm}
\end{figure}
The experimental setup is shown in Fig.\ref{fig:ExperimentalSetup1}, where two possible end-effectors are considered. Specifically, since the initial goal of the experiments is to evaluate the impact of control parameters on task execution—particularly the stiffness level $\chi$, the channel radius $\delta$, and the implementation of additional assisting or opposing mechanisms—a simple handle has been attached to the force/torque sensor, as shown in Fig.\ref{fig:ExperimentalSetup1}(a). In the second phase of the experiments, to test the proposed architecture during the assisted execution of Activities of Daily Living (ADLs) \cite{lauretti2017learning}, a purpose-designed constraining mechanism for the patient's wrist was installed on the robot, as shown in Fig.\ref{fig:ExperimentalSetup1}(b).
}
\vspace{-5mm}
\subsection{Path Generation via LbD}
\begin{figure}[tb]
    \centering
    \psfrag{x}[c][c][0.75][0]{ $x$ {\small [m]}}
    \psfrag{z}[c][c][0.75][0]{ $z$ {\small [m]}}
\includegraphics[width=0.49\columnwidth]{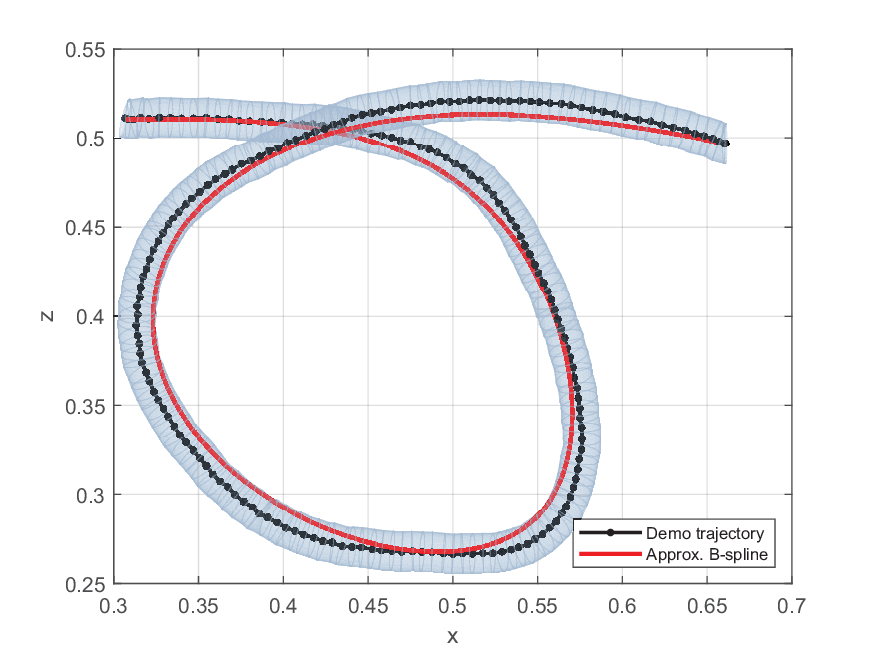} %C:\Users\admin\OneDrive - Unimore\Documenti\MATLAB\LbD_ConstrainedSmoothingBsplineApproximation\TubeApproximateFreeFormConstantVariableBound1.m
\includegraphics[width=0.49\columnwidth]{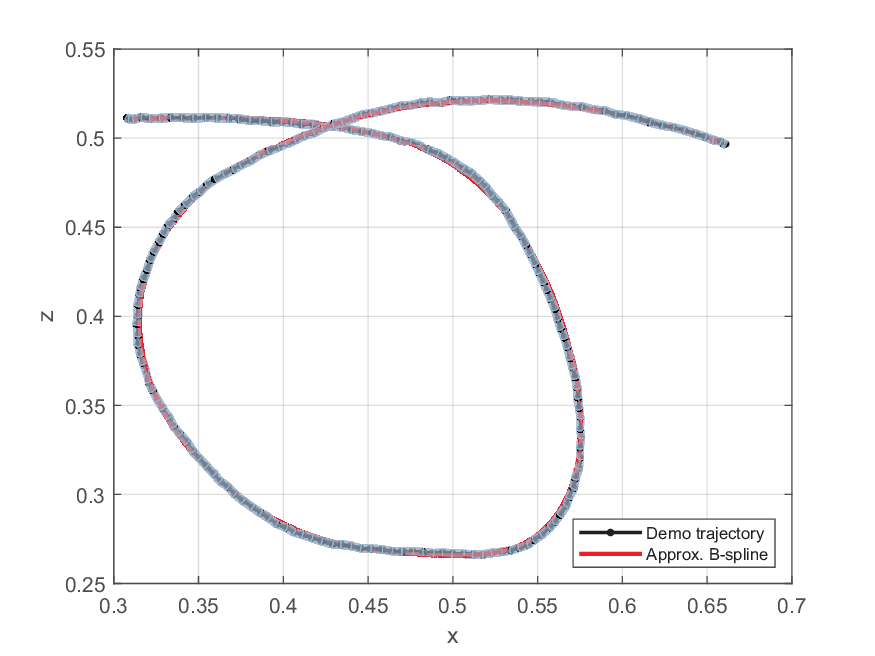}\\[-1mm]
{\small (a) \hspace{40mm}(b)}\\[-1mm]
    %%%Versione di Luigi
    % \caption{\black{Approximation of the demonstrated trajectory using a constrained smoothing B-spline with $\lambda = 0.01$ and a maximum tolerance of $\varepsilon = 0.01$ [m] (a) and $\varepsilon = 0.002$ [m] (b), respectively.}}
    %%%Fine version di Luigi
    \caption{\black{Approximation of the demonstrated trajectory using a constrained smoothing B-spline with $\lambda = 0.01$ and a maximum tolerance of $\varepsilon = 0.01$ [m] (a) and $\varepsilon = 0.002$ [m] (b), respectively. The gray zone represents the feasible region within the prescribed tolerance  $\varepsilon$ from the recorded data points.}}
    \label{fig:Reference path}
    \vspace{-5mm}
\end{figure}

In order to define the geometric path for an exercise, the therapist guides the robot's end-effector, which in this case is a simple handle, within its operational range and the sequence of points, sampled with period $T_s$, are recorded. These points are then interpolated with a B-spline curve, i.e. a parametric curve $\boldphi(u):[u_{\min}, u_{\max}] \rightarrow\mathbb{R}^3$  defined as  linear combinations of {\it control points} $\p_j \in \mathbb{R}^3$ weighted by {\it B-spline basis functions} of degree $p$, $B_j^p(u)$:
  \be%
\boldphi(u)= \sum_{j=0}^{N} \p_j B_j^p(u),~~~~~~~u_{\min} \le u
 \le u_{\max}.
\label{eq:BsplineDef}
\ee
 The vectorial coefficients $\p_j, \, j=0,\ldots,\,N$, determine the shape of the curve and are computed by imposing
 approximation conditions on the samples ${\bf q}_j$, $j=0,\ldots,n$ of the recorded trajectory. In particular, to suppress unwanted movements that affect the user's motion, {\it smoothing} B-splines  are considered since they minimize the cost function:
 \begin{equation}
J:=\sum_{j=0}^{n}w_j\|\boldphi(u^\star_j)-{\bf q}_j\|^2 +\lambda \int_{u_{\min}}^{u_{\max}}\left\|
\frac{d^2 \boldphi(\tau)}{d \tau^2}\right\|^2d\tau. \label{eq:CostFunction}\end{equation}
Therefore, they represent a trade-off between the squared approximation error with respect to the demonstrated trajectory and the smoothness of the resulting curve.
The parameter  $\lambda \ge 0$ can be freely chosen
to govern this trade-off, while $w_j>0$ is a parameter used to selectively weight the contribution of the squared error at a particular point  ${\bf q}_j$.
\black{Since the selection of individual weights $w_j$ is not feasible for general applications without human intervention and a trial-and-error procedure, and finding the right trade-off between smoothness and approximation precision by adjusting only the parameter $\lambda$ may prove impossible, a constrained approach for defining $\boldphi(u)$ has been adopted, as suggested in \cite{Biagiotti2023}.
Accordingly, the control points $\p_j$ in (\ref{eq:BsplineDef}) are computed by minimizing the cost function $J$ in (\ref{eq:CostFunction}), subject to the constraint:
\[
\|\boldphi(u^\star_j)-{\bf q}_j\|\le \varepsilon
\]
where $\varepsilon$ is a scalar value.
This approach allows imposing the desired tolerance $\varepsilon$ between the data points recorded from the demonstrated trajectory and the resulting approximating function $\boldphi(u)$. In \Fig{Reference path}, the B-spline functions obtained from the same demonstrated trajectory with two different values of $\varepsilon$ are shown.  For clarity, a planar motion in the $x$-$z$ plane has been considered.
}
It is worth noting that the path exhibits an intersection at a certain point. Consequently, while there is no uncertainty regarding the position along the path, the normal direction at this point is not uniquely defined.
% As a final note, it is worth highlighting that in these experiments, the orientation of the end-effector has been kept constant,aligned with the axes of the base reference frame. However, despite the mechanism for Human-Robot interaction being solely based on position, being the variable $s$ the arc-length along the spatial curve $\boldphi$, it is possible to define a function $\boldphi_o(s)$ that provides a minimal representation of the orientation as a function of progress along the path. This curve can be obtained by interpolating the orientations imposed by the therapist during the demonstration, just like the position.
As a final note, it is worth highlighting that in these experiments, the orientation of the end-effector was kept constant and aligned with the axes of the base reference frame. However, although the mechanism for Human-Robot interaction is solely based on position—where the variable \( s \) represents the arc-length along the spatial curve \( \boldphi \)—it is possible to define a function \( \boldphi_o(s) \) that provides a minimal representation of the orientation as a function of progress along the path. This curve can be obtained by interpolating the orientations imposed by the therapist during the demonstration, in the same way as the position.
 \vspace{-2.5mm}
\black{\section{Experimental results and discussion}
\label{Sec.Experimental_results}
To validate the proposed control architecture and evaluate the impact of the control parameters on system performance, we conducted several tests under various control conditions. These tests emulated a rehabilitation exercise along the predefined path shown in Fig.~\ref{fig:Reference path}. During these tests, we collected the forces exchanged during Human-Robot Interaction and measured the user's deviation from the reference path.\\
\begin{figure}[tb]
    \centering
    \begin{minipage}[b]{0.33\columnwidth}
        \centering
         \footnotesize{\hspace{1 cm} $\delta = 0.01 [m]$}
    \end{minipage}%
    \begin{minipage}[b]{0.33\columnwidth}
        \centering
        \footnotesize{\hspace{1 cm} $\delta = 0.02 [m]$}
    \end{minipage}%
    \begin{minipage}[b]{0.33\columnwidth}
        \centering
        \footnotesize{\hspace{1 cm} $\delta = 0.03 [m]$}
    \end{minipage}\\%[1ex] % First row
    \begin{minipage}[c]{0.04\textwidth}
        \centering
        \footnotesize{\rotatebox{90}{$\chi = 100 [N/m]$}}
    \end{minipage}%
    \begin{minipage}[c]{0.33\columnwidth}
        \centering
        \psfrag{X}[c][c][0.8][0]{X [m]}
        \psfrag{Y}[c][c][0.8][0]{Y [m]}
        \psfrag{Z}[c][c][0.8][0]{Z [m]}
        \includegraphics[width=\columnwidth]{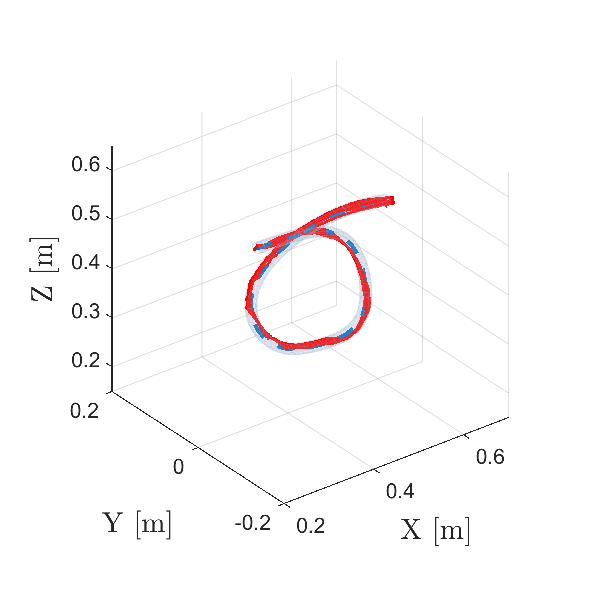}
    \end{minipage}%
    \begin{minipage}[c]{0.33\columnwidth}
        \centering
        \psfrag{X}[c][c][0.8][0]{X [m]}
        \psfrag{Y}[c][c][0.8][0]{Y [m]}
        \psfrag{Z}[c][c][0.8][0]{Z [m]}
        \includegraphics[width=\columnwidth]{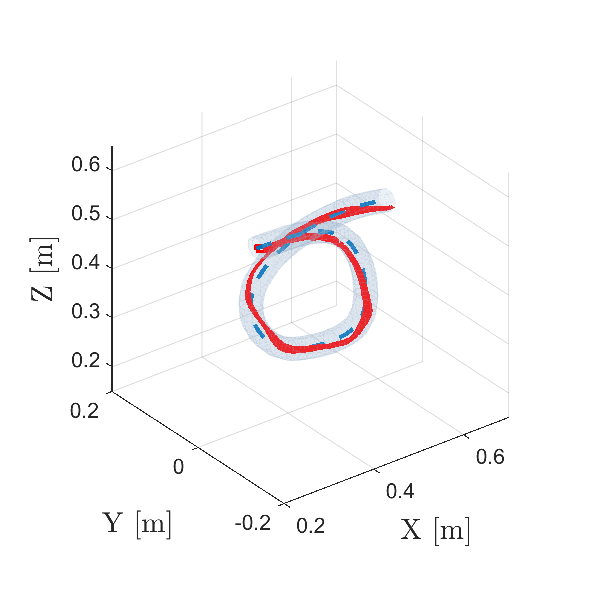}
    \end{minipage}%
    \begin{minipage}[c]{0.33\columnwidth}
        \centering
        \psfrag{X}[c][c][0.8][0]{X [m]}
        \psfrag{Y}[c][c][0.8][0]{Y [m]}
        \psfrag{Z}[c][c][0.8][0]{Z [m]}
        \includegraphics[width=\columnwidth]{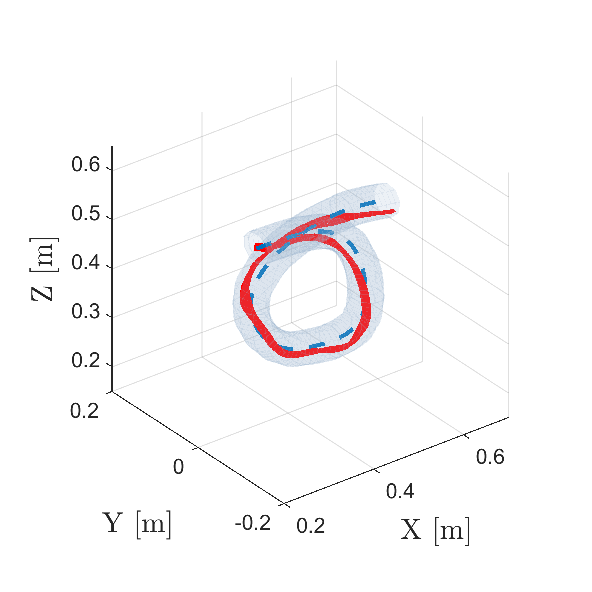}
    \end{minipage}\\%[2ex]     % Second row
    \begin{minipage}[c]{0.04\textwidth}
        \centering
        \footnotesize{\rotatebox{90}{$\chi = 500[N/m]$}}
    \end{minipage}%
    \begin{minipage}[c]{0.33\columnwidth}
        \centering
        \psfrag{X}[c][c][0.8][0]{X [m]}
        \psfrag{Y}[c][c][0.8][0]{Y [m]}
        \psfrag{Z}[c][c][0.8][0]{Z [m]}
        \includegraphics[width=\columnwidth]{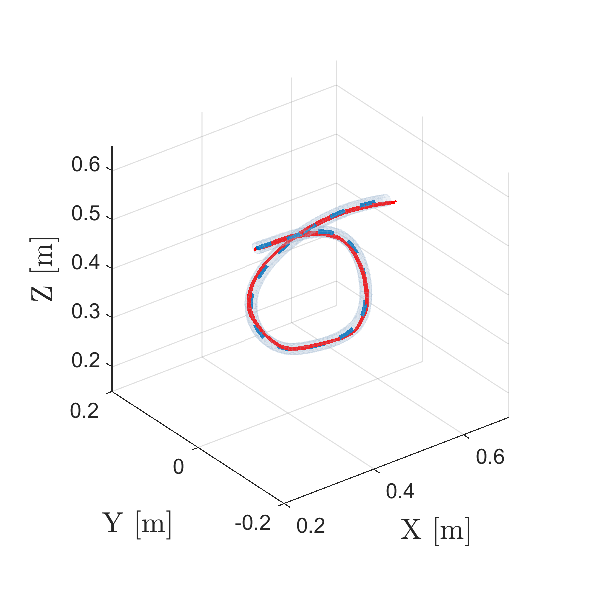}
    \end{minipage}%
    \begin{minipage}[c]{0.33\columnwidth}
        \centering
        \psfrag{X}[c][c][0.8][0]{X [m]}
        \psfrag{Y}[c][c][0.8][0]{Y [m]}
        \psfrag{Z}[c][c][0.8][0]{Z [m]}
        \includegraphics[width=\columnwidth]{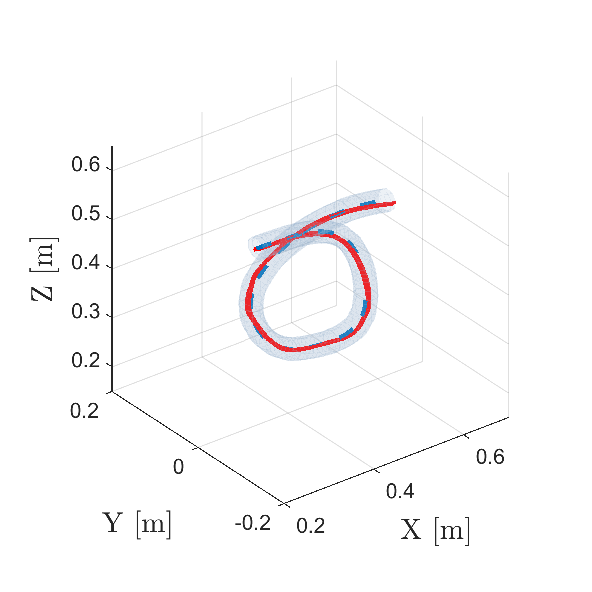}
    \end{minipage}%
    \begin{minipage}[c]{0.33\columnwidth}
        \centering
        \psfrag{X}[c][c][0.8][0]{X [m]}
        \psfrag{Y}[c][c][0.8][0]{Y [m]}
        \psfrag{Z}[c][c][0.8][0]{Z [m]}
        \includegraphics[width=\columnwidth]{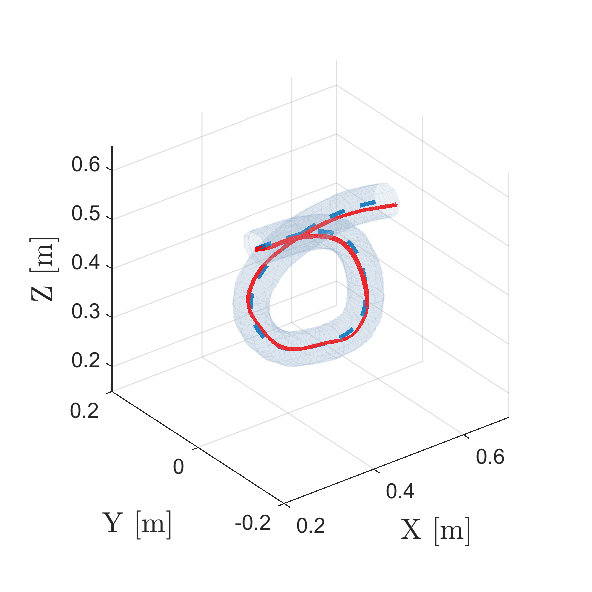}
    \end{minipage}\\%[2ex]        % Third row
    \begin{minipage}[c]{0.04\textwidth}
        \centering
        \footnotesize{\rotatebox{90}{$\chi = 2500[N/m]$}}
    \end{minipage}%
    \begin{minipage}[c]{0.33\columnwidth}
        \centering
        \psfrag{X}[c][c][0.8][0]{X [m]}
        \psfrag{Y}[c][c][0.8][0]{Y [m]}
        \psfrag{Z}[c][c][0.8][0]{Z [m]}
        \includegraphics[width=\columnwidth]{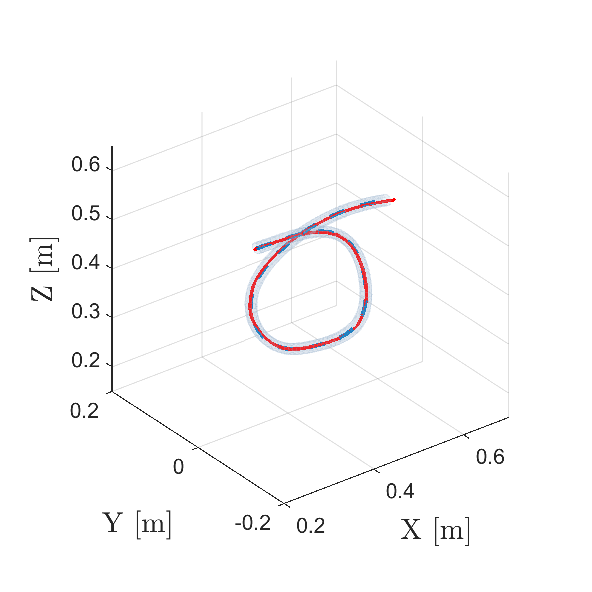}
    \end{minipage}%
    \begin{minipage}[c]{0.33\columnwidth}
        \centering
        \psfrag{X}[c][c][0.8][0]{X [m]}
        \psfrag{Y}[c][c][0.8][0]{Y [m]}
        \psfrag{Z}[c][c][0.8][0]{Z [m]}
        \includegraphics[width=\columnwidth]{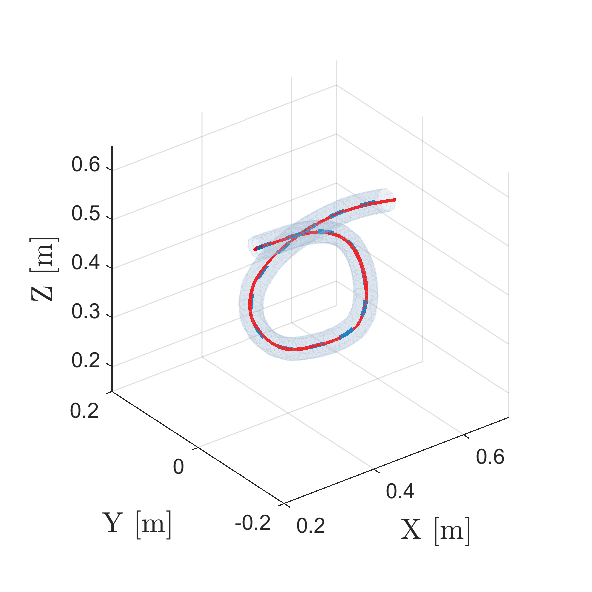}
    \end{minipage}%
    \begin{minipage}[c]{0.33\columnwidth}
        \centering
        \psfrag{X}[c][c][0.8][0]{X [m]}
        \psfrag{Y}[c][c][0.8][0]{Y [m]}
        \psfrag{Z}[c][c][0.8][0]{Z [m]}
        \includegraphics[width=\columnwidth]{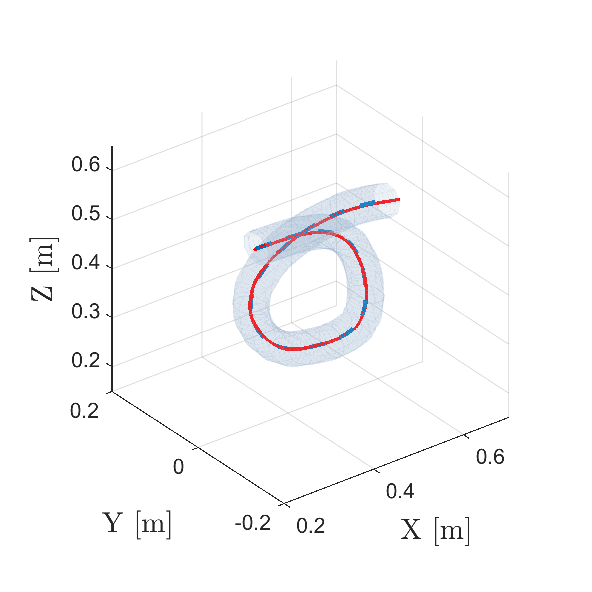}
    \end{minipage}\\
    \caption{Executions of the rehabilitation task defined in \Fig{Reference path} recorded under different values for $\chi$ and $\delta$.}
    \label{fig:9_tests}
\end{figure}
 \vspace{-5mm}
 \subsection{Technical Validation of the Control Parameters}

 In the initial experiments, a healthy user with experience conducted tests to analyse how different control parameters influenced the system’s performance. We first focused on the parameters\footnote{The three values considered are representative of low, medium, and high levels.} $\chi \in \{100, 500, 2500\} \, \text{[N/m]}$ and $\delta \in \{0.01, 0.02, 0.03\} \, \text{[m]}$, resulting in 3 × 3 different scenarios depicted in Fig.~\ref{fig:9_tests}.

 For all these tests, mass dynamics parameters were fixed at $m = 5 \, \text{[kg]}$ and $b = 15 \, \text{[Ns/m]}$. As $\chi$ increased, users experienced stronger guiding forces toward the center of the channel, helping them stay aligned with the path. Conversely, increasing $\delta$ while keeping $\chi$ constant weakened the guiding force, allowing more lateral movement before users encountered resistance, creating a “wall” effect at the channel boundaries.
\vspace{-5mm}
\subsection{Trade-Off Between Stiffness ($\chi$) and Radius ($\delta$)}
Figure~\ref{fig:E_N subplot} illustrates the deviation magnitude from the reference path in the orthogonal direction ($\| \tilde{x}_{\perp} \|$) for each $(\chi, \delta)$ combination. The results indicate that higher $\chi$ values enhance path-following precision, while $\delta$ affects the maximum permissible lateral displacement. However, for high $\chi$ values, $\delta$ has minimal impact on the average deviation.

%\textcolor{black}{In this study, we assessed the effectiveness of the constraint in guiding the user's task execution. The perpendicular component of the interaction force, $\mathbf{F}_{\perp}$, serves as a crucial indicator of the patient's ability to track the reference trajectory accurately. When $\mathbf{F}_{\perp}$ remains close to zero, it signifies precise adherence to the path, indicating that the patient may be capable of executing the task independently.}

\textcolor{black}{ By comparing these results on the tracking errors with the normal forces exerted by the user during the motion, shown in Figure~\ref{fig:F_N subplot}, it clearly emerges the importance of haptic cues for the human to stay close to the desired path. In fact, variations in $\delta$ do not significantly influence these forces, while increasing $\chi$ slightly raises the average force, which in any case remains about $5$–$6$ N throughout the experiment. This suggests that this is the minimum level of force necessary to guide a healthy and collaborative\footnote{The task required of the subject involved in the experiments was to track the virtual path.} user along the path. As a consequence of the different parameter values, this level of force is then translated into a greater or lesser distance from the reference curve. Interestingly, if the peak forces applied in each experiment (Fig.~\ref{fig:NEW_F_N subplot}) are analyzed, they reveal that lower values of $\chi$ result in higher peak forces. This phenomenon is particularly pronounced for smaller $\delta$ values, where users tend to "bounce" between the virtual channel walls, leading to sudden force spikes.
In conclusion, the stiffness $\chi$ and radius $\delta$  mainly affect the user's ability to accurately reproduce the planned motion but not the exchanged forces. High levels of $\chi$ tend to smoothly guide the user without producing jerky behaviors, which are instead caused by small radii $\delta$  in conjunction with low levels of stiffness $\chi$. When used with unhealthy subjects (not tested in the present study), sufficiently large values of $\delta$  should be considered to allow the limb with limited mobility to deviate from the prescribed path without excessive forces. At the same time, an adequate level of $\chi$ should be imposed to guide the user in performing the correct movements.
}

\begin{figure}[bt]
    \centering
    \includegraphics[width=\columnwidth]{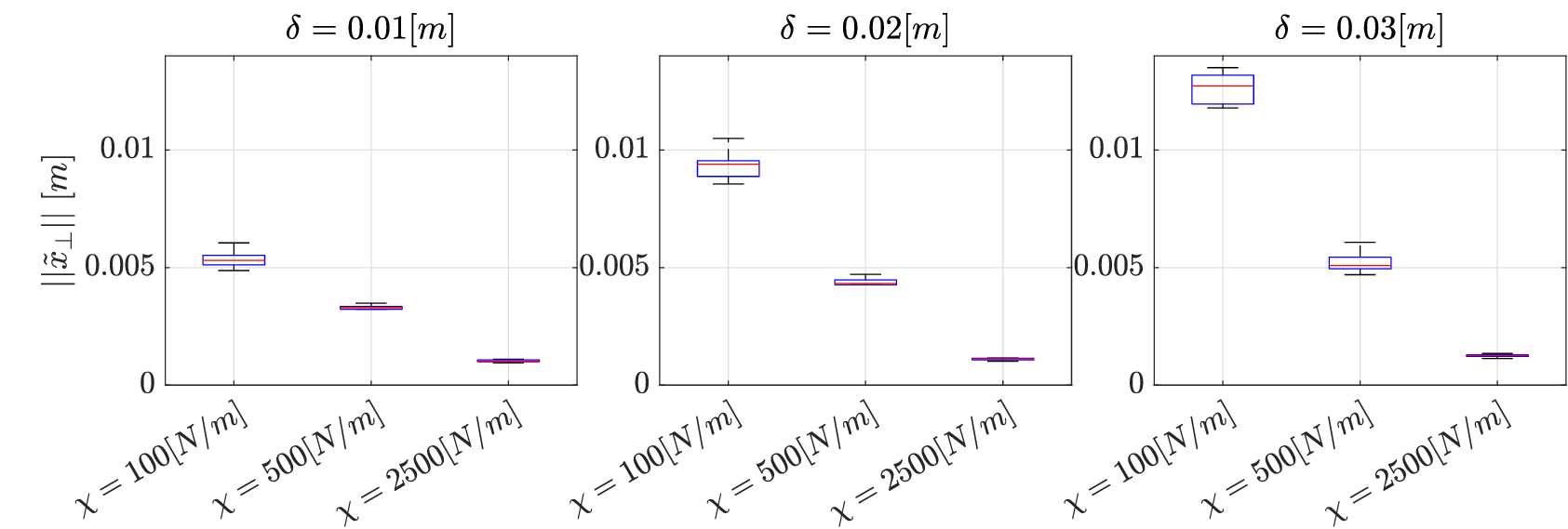}
    \caption{Deviation with respect to the reference path in the orthogonal direction, obtained in the experiments shown in Fig.\ref{fig:9_tests} for different values of parameters $\chi$ and $\delta$.}
    \label{fig:E_N subplot}
        \vspace{-5mm}
\end{figure}

\begin{figure}[bt]
    \centering
    \includegraphics[clip,width=\columnwidth]{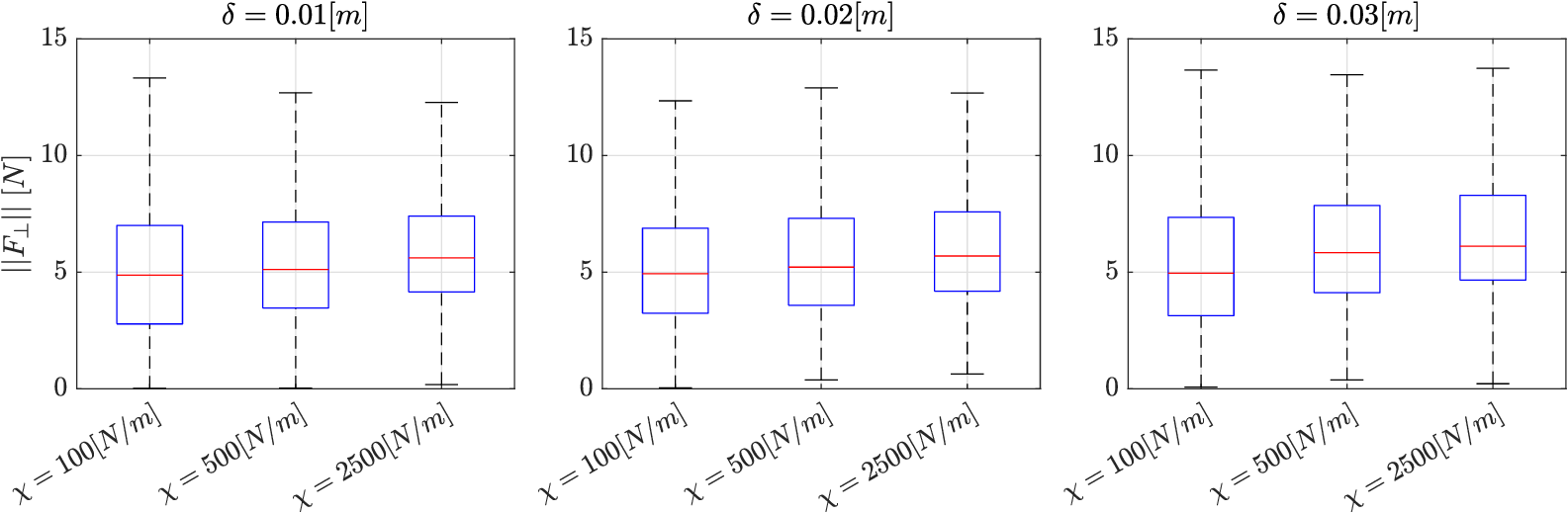}
    \caption{Normal forces exchanged in the experiments shown in Fig.\ref{fig:9_tests} for different values of parameters $\chi$ and $\delta$.}
    \label{fig:F_N subplot}
        \vspace{-5mm}
\end{figure}

%Figure~\ref{fig:F_N subplot} shows the normal forces exchanged between the robot and the user to maintain the desired path. \textcolor{black}{The results indicate that variations in $\delta$ do not significantly influence these forces, while increasing $\chi$ slightly raises the average force. However, further analysis of the peak forces in each experiment (Fig.~\ref{fig:NEW_F_N subplot}) reveals that lower values of $\chi$ result in higher peak forces. This phenomenon is particularly pronounced for smaller $\delta$ values, where users tend to "bounce" between the virtual channel walls, leading to sudden force spikes.}
\begin{figure}[bt]
    \centering
    \includegraphics[clip,width=\columnwidth]{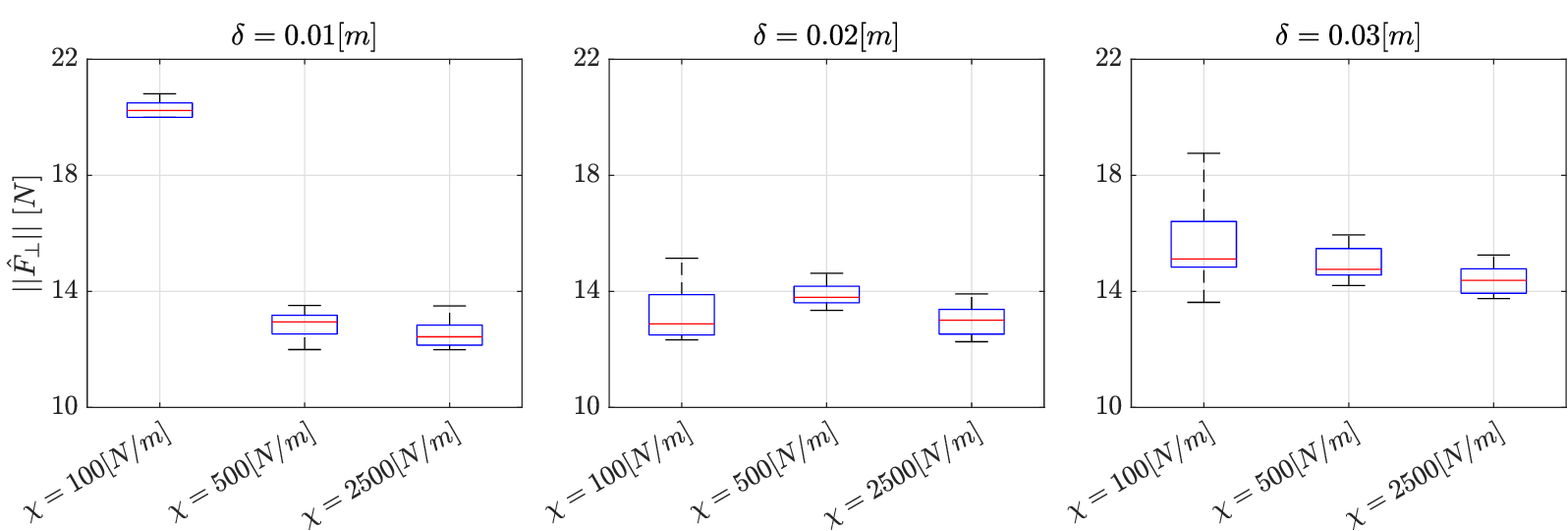}
    \caption{Peak values of the normal forces exchanged in the experiments shown in Fig.\ref{fig:9_tests} for different values of parameters $\chi$ and $\delta$.}
    \label{fig:NEW_F_N subplot}
        \vspace{-5mm}
\end{figure}

%\textcolor{black}{Comparing these findings, collected in Figures~\ref{fig:F_N subplot} and~\ref{fig:NEW_F_N subplot}, which show the perpendicular forces and their peak values under various operating conditions, with the results presented in Figure~\ref{fig:E_N subplot}, reveals that variations in channel diameter~$\delta$ and stiffness~$\chi$ significantly affect the maximum allowable execution error. However, these changes do not negatively impact the user's perception or the practical usability of the system in real-world scenarios. This is because the forces exchanged between the robot and the user, specifically the perpendicular forces along the curve, remain consistent across different constraint conditions, whether the guidance is more lenient or more restrictive. Such stability highlights the robustness of the proposed constraint approach, ensuring uniform user feedback and maintaining reliable and predictable interaction during task execution.}

%%%%%%%%%%end_second_second_review%%%%%%%
\vspace{-4mm}
 \subsection{Influence of Virtual Mass Parameters ($m$, $b$)}
 \label{subsec:virtual_mass_influence}
Additionally, we examined the impact of the virtual mass parameters $m$ and $b$ on system dynamics, \textcolor{black}{emphasizing their role in defining human-robot interaction through the parallel force component ($\mathbf{F}_{\parallel}$) during task execution.} As shown in Fig.~\ref{fig:F_tau subplot}, reducing the parameters to $m = 1 \, \text{[kg]}$ and $b = 3 \, \text{[Ns/m]}$ decreased the user-applied tangential force component, \textcolor{black}{thereby lowering the physical effort required for task completion. This result highlights how tuning $m$ and $b$ effectively adapts the virtual mass dynamics to accommodate different user profiles and physical conditions, directly impacting the practical usability of the system by reducing fatigue and improving comfort during prolonged use.}

\textcolor{black}{Furthermore, Figures~\ref{fig:F_tau subplot} and~\ref{fig:Assit_Oppos_Force} illustrate that the parallel force component remains independent of the constraint parameters and is influenced solely by changes in virtual mass parameters or additional assistive or resistive forces. In particular, Fig.~\ref{fig:Assit_Oppos_Force} demonstrates how the introduction of a $\pm 1 \, \text{N}$ virtual force can provide either assistive (positive) or resistive (negative) support, directly modifying the tangential forces exerted by the user. This feature offers a customizable assistive mechanism, allowing the system to dynamically adapt to the user's capabilities and needs.}

\textcolor{black}{This result underscores the system’s high modularity and adaptability, enabling adjustments to task complexity and assistance levels without compromising the integrity of the constraint. By allowing precise tuning of $m$ and $b$, the system ensures optimal practical usability across a range of user profiles and rehabilitation goals. This flexibility supports effective assistance under various operating conditions and facilitates personalized rehabilitation pathways tailored to patient progress and recovery needs.}

\begin{figure}[t]
    \centering
    \includegraphics[width=0.8\columnwidth,height=0.32\columnwidth]{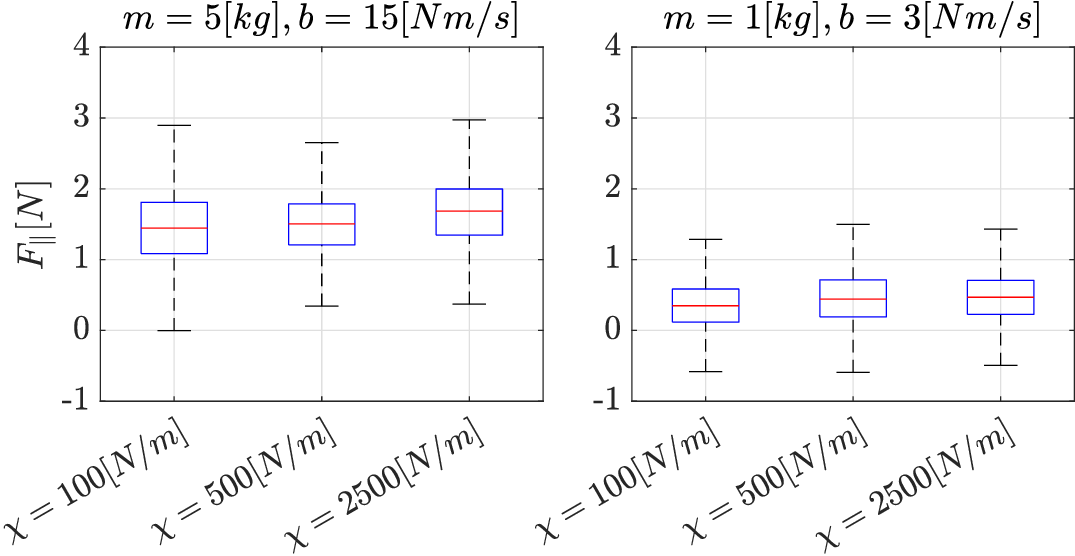}
    \caption{Tangential forces exerted by the user with different values of $m$ and $b$ ($\delta = 0.03 \; \text{m}$).}
    \label{fig:F_tau subplot}
    \vspace{-5mm}
\end{figure}

\begin{figure}[tb]
    \centering
    \includegraphics[width=0.8\columnwidth,height=0.32\columnwidth]{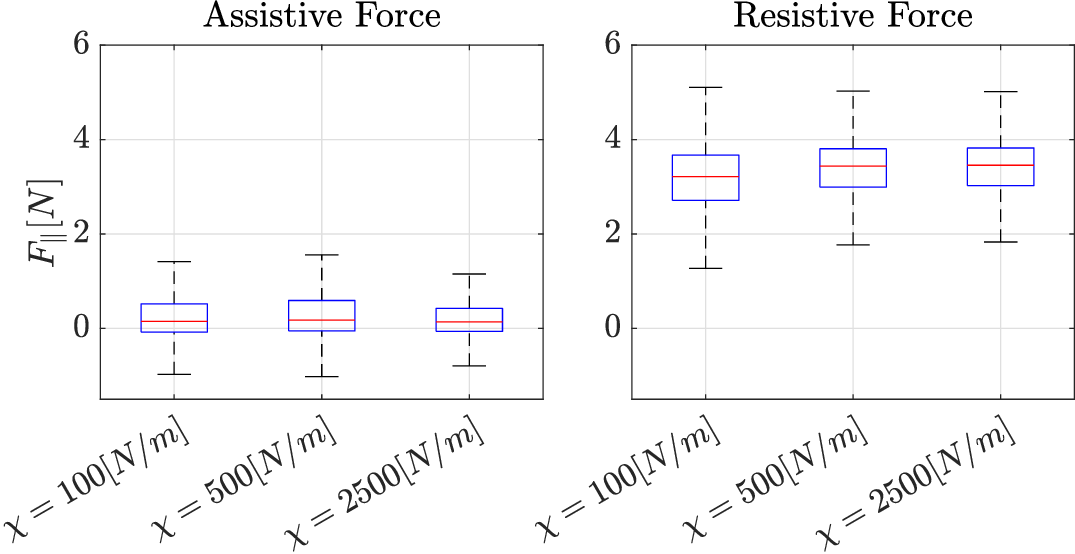}
    \caption{Tangential forces exerted by the user with the application of $1$ N assistive or resistive force.}
    \label{fig:Assit_Oppos_Force}
    \vspace{-5mm}
\end{figure}

%%%%%END_Second_review%%%%%
\vspace{-4mm}
 \subsection{Statistical Analysis of User Perception and Evaluation}
To further validate the system, we evaluated user perceptions of task execution and interaction with the robot. We recruited 10 healthy participants (7 males, 3 females, aged 24 to 65 years, mean age 32 years) with no prior experience with robots. %\textcolor{black}{All participants received detailed information and provided informed consent before starting the experiments. The study was approved by the Ethics Committee of the University of Modena and Reggio Emilia.}

A concise questionnaire, adapted from the NASA-TLX (Task Load Index), was administered to assess user experience under different control parameter configurations (see Fig.~\ref{fig:NEW_questionnaire}). Participants rated statements on \textit{Physical Demand, Performance, Effort,} and \textit{Frustration} using a 6-point scale (1 = Very low, 6 = Very high). Six conditions from Fig.~\ref{fig:9_tests} were selected, starting with initial settings of $\chi = 100$ [N/m] and $\delta = 0.01$ [m], and incrementally increasing $\chi$ and $\delta$.\\

To complement the evaluation, we performed a one-way ANOVA to analyze subjective ratings across conditions, applying a significance threshold of $p < 0.05$. Post-hoc Tukey tests were conducted for significant results.

\begin{itemize}
    \item \textit{Physical Demand ($p = 0.7547$):} No significant differences were observed across conditions.
    \item \textit{Performance ($p = 0.000009$):} Significant differences emerged, particularly with higher $\chi$. Group 1 ($\chi = 100$ N/m, $\delta = 0.01$ m) differed notably, indicating improved performance with increased $\chi$.
    \item \textit{Effort ($p = 0.0055$):} Higher $\chi$ values reduced perceived effort. Group 3 ($\chi = 2500$ N/m, $\delta = 0.01$ m) was rated as less effortful than Group 1.
    \item \textit{Frustration ($p = 0.0084$):} No significant differences, indicating $\chi$ and $\delta$ did not strongly influence frustration.
\end{itemize}

\begin{figure}[tb]
    \centering
    \includegraphics[width=\columnwidth,height=0.7\columnwidth]{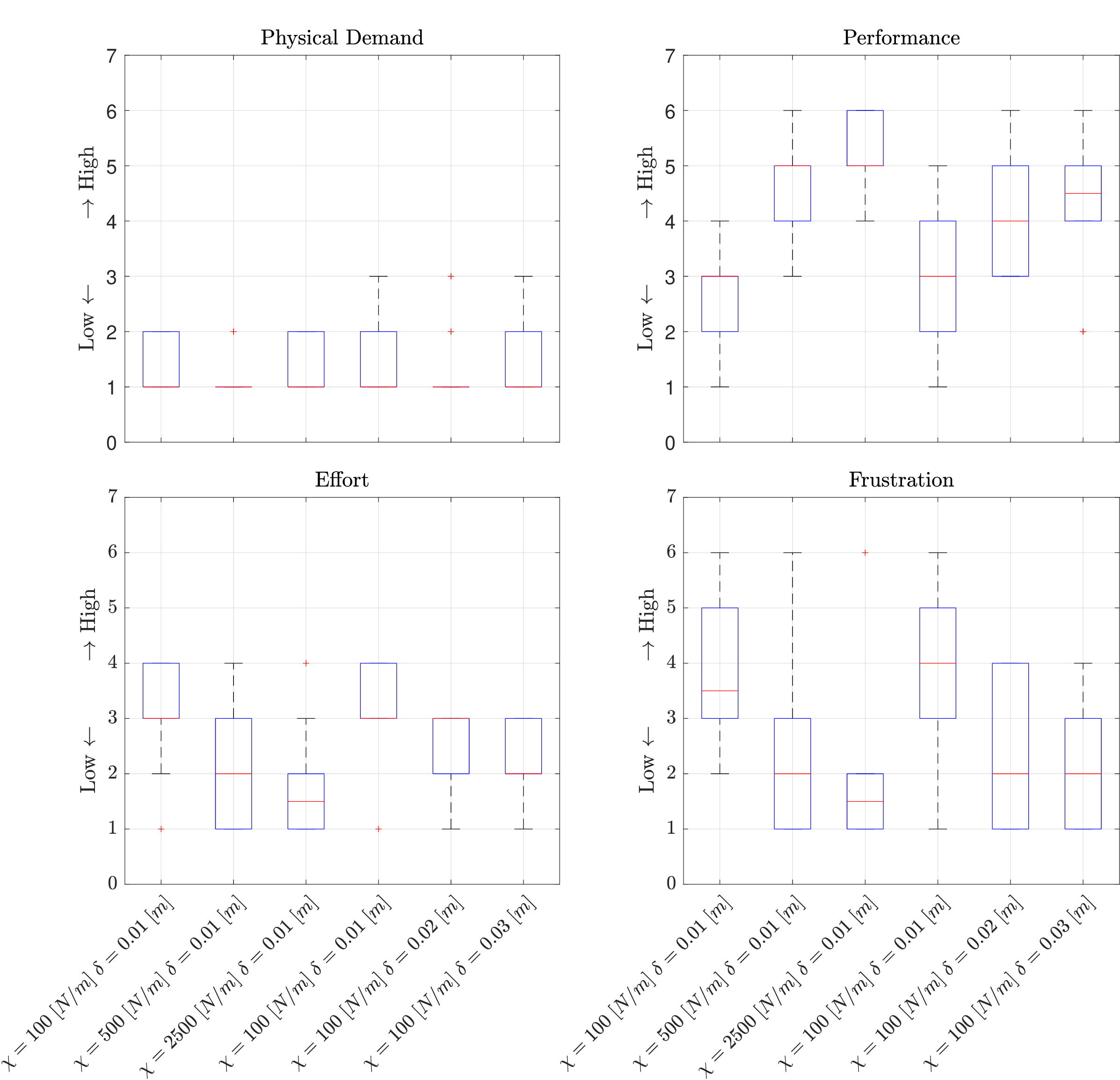}
    \caption{Users' opinions about Physical Demand, Performance, Effort, and Frustration related to the tests in Fig.~\ref{fig:9_tests}.}
    \label{fig:NEW_questionnaire}
    \vspace{-5mm}
\end{figure}
 These findings corroborate the technical results, confirming that higher \textit{Stiffness ($\chi$)} values improved perceived performance quality, as users reported fewer execution errors and felt more efficient following the path. Higher $\chi$ also reduced perceived effort, as the stronger guidance toward the path decreased the effort required. In particular, as highlighted in the previous subsection~\ref{subsec:virtual_mass_influence}, \textit{Physical Demand} did not show any variation with changes in $\chi$ or $\delta$ because it is strictly dependent on the combination of the virtual mass dynamics parameters $m$ and $b$.

 Furthermore, increasing the \textit{Channel Radius ($\delta$)} allowed greater freedom in task execution, reducing frustration as users felt less constrained. Users experienced improved performance due to more manoeuvring space within the channel boundaries, especially at higher $\chi$ values.

\textcolor{black}{Despite the limited sample size, %, the lack of sample diversity, and the use of only subjective measurements,
the observed trends are well-defined and statistically supported, as confirmed by the results of the ANOVA and post-hoc Tukey tests. The patterns are consistent across participants and exhibit minimal variability, reinforcing their reliability. Additionally, no anomalies were detected, further validating the robustness of the trends.% Notably, the statistical significance achieved, even with a small sample, highlights the strength of the observed effects and underscores the clarity and consistency of the results obtained.
}

%%%%%%%%end_second_review%%%%%%%%%%%
\vspace{-5mm}
\subsection{Performance Comparison}
}
To evaluate the advantages of our proposed solution over a widely used method in the rehabilitation field, we conducted a comparative study. Specifically, we compared our method with the approach presented in \cite{lauretti2017learning}, which is based on the Dynamic Movement Primitives (DMP) formulation and applied to a typical rehabilitation task for daily living activities. This approach was selected because it explicitly utilizes a Learning by Demonstration procedure to define the rehabilitation task, which is one of the main features of our method. In this scenario, the robot assists the patient in reaching and picking up a bottle from a table, as illustrated in Fig.~\ref{fig:ExperimentalSetup1}(b).

For this study, a custom-designed end-effector was developed to securely connect the user's wrist to the robot's terminal flange, ensuring both comfort and effective interaction. This specialized end-effector facilitates precise task execution and enables efficient force transfer between the user and the manipulator during rehabilitation exercises.

After demonstrating the task by guiding the robot under gravity-compensation conditions, both algorithms were applied, and the results are presented in the following figures.

\begin{figure}[tb]
\centering
\includegraphics[width=\columnwidth,height=0.7\columnwidth]{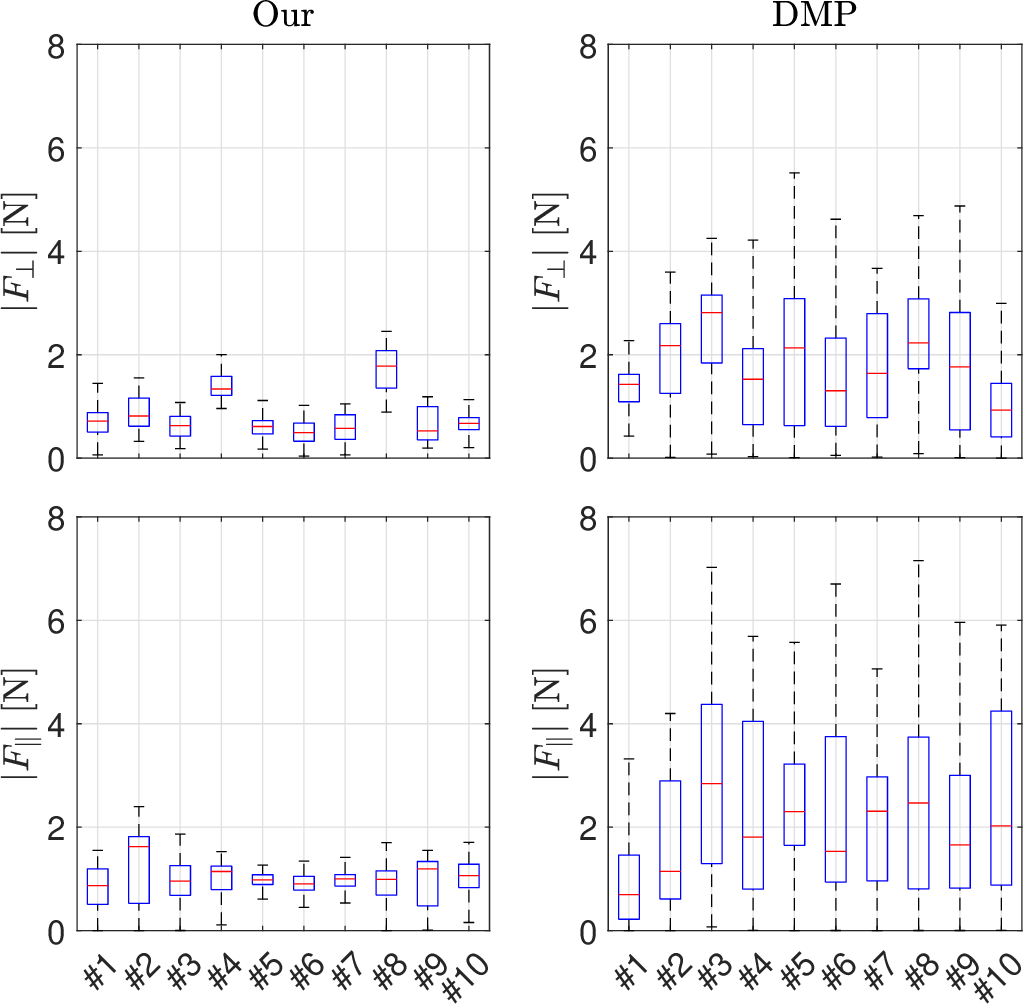}
\caption{Forces recorded during the comparison between our method and the DMP-based approach.}
\label{fig:boxplot_DMP}
    \vspace{-5mm}
\end{figure}

As shown in Fig.~\ref{fig:boxplot_DMP}, the forces exerted during task execution using the method proposed in \cite{lauretti2017learning} were generally higher than those recorded with our method. The force magnitudes are plotted to emphasize their relative intensities.

An additional key observation relates to the nature of patient interaction. In the DMP-based approach, the patient operates exclusively in a passive mode, often resulting in forces being applied in the opposite direction to the desired motion.
\textcolor{black}{By contrast, our method enables a broader range of rehabilitation modes, from fully passive assistance in the early stages to more active participation as the patient’s abilities progress, thereby eliminating the rigid passivity of DMP-based systems.}

\textcolor{black}{
The comparison between our \textit{Admittance Guidance Virtual Fixture} control and the DMP-based method highlights several key advantages that have direct implications for real-world rehabilitation applications:
}
\textcolor{black}{
\begin{itemize}
    \item \textbf{Enhanced patient comfort:} Our approach significantly reduces interaction forces compared to the DMP-based method. This is due to the adaptive compliance of the virtual fixture, which accommodates natural variations in patient movement without generating excessive corrective forces, thereby enhancing user comfort and reducing fatigue.
    \item \textbf{Active user engagement:} Unlike the DMP-based approach, which enforces predefined movement patterns \textcolor{black}{(and remains limited to passive rehabilitation)}, our method allows controlled deviations within a virtual path. This fosters greater patient participation, making rehabilitation more engaging and motivating, which is crucial for long-term adherence.
    \item \textbf{Improved adaptability:} The stiffness level ($\chi$) and channel width ($\delta$) can be dynamically adjusted to match different rehabilitation phases, allowing a smooth transition from guided exercises in early rehabilitation to more autonomous movements in later stages.
\end{itemize}
}
\textcolor{black}{
A crucial distinction of our approach is its flexibility in trajectory execution. Unlike DMP-based methods, which impose rigid timing constraints, our system allows patients to regulate their movement speed autonomously, supporting a more natural rehabilitation process. This adaptability facilitates both passive and active rehabilitation scenarios, adjusting the level of assistance based on the patient’s progress.\\
}
\textcolor{black}{
Moreover, our system enables further customization by allowing the application of assistive or resistive forces, tailoring the rehabilitation experience to the individual’s specific needs. This ability to dynamically modulate task difficulty contributes to a patient-centered approach that enhances both comfort and effectiveness in rehabilitation.\\
}
\textcolor{black}{
These findings highlight that our method is particularly advantageous for progressive rehabilitation, where the required assistance level evolves over time. By offering greater flexibility in guidance and effort modulation, our approach supports a structured transition from externally guided movements to self-initiated motor execution. This structured adaptation is essential for maximizing motor learning during the rehabilitation process.\\
}

 \vspace{-1cm}
\black{\section{Conclusions}}
\label{sec:Conclusions_real}
This paper presents a layout optimization method and a novel control architecture for collaborative robots applied to upper limb rehabilitation. The system offers therapist-friendly programmability through Programming by Demonstration (PbD) and incorporates an admittance-type Virtual Fixture control.\footnote{Watch a video explaining the procedure at this link: \url{https://youtu.be/mTbwHzUQS_g}.} This control strategy allows the robot to act as a supportive guide for the patient's movements. \textcolor{black}{A key objective of our study was to validate the system’s applicability within upper-limb rehabilitation, demonstrating its potential to replace multiple tools with a single, versatile solution.}

A unique feature of this approach is the combination of passive and active control modes within the framework of admittance control. By leveraging virtual fixtures, the robot can provide tailored resistance or support, adapting exercise difficulty to the patient’s progress. These capabilities are further enhanced by an optimization methodology that maximizes the robot's effectiveness in the workspace, considering payload constraints and manipulability indices.

Experimental validation involved the execution of tasks generated using a constrained smoothing B-spline approach, enhanced by the Virtual Fixture mechanism. The results demonstrated the method's robustness across varying control parameters, including stiffness ($\chi$) and channel radius ($\delta$). Higher stiffness values improved trajectory precision, while the channel radius determined the allowable deviations, enabling flexible guidance. Statistical analysis from user studies further revealed that these parameters significantly influenced perceived performance and effort, aligning well with the technical results.

\textcolor{black}{The proposed system introduces significant technical innovations along with several practical benefits. It offers dynamic adaptation to patient needs by modulating assistance levels in real-time, supports personalized assistance/resistance strategies, and ensures ease of integration into clinical environments through a Learning by Demonstration approach that simplifies task programming. Additionally, by adjusting key parameters ($m$, $b$, $\chi$, and $\delta$), the system effectively modulates task difficulty to progressively challenge patients. %While our current comparative analysis confirms its effectiveness in reducing undesired forces, further quantitative and qualitative comparisons with existing systems could strengthen the overall evaluation, paving the way for enhanced applications in robotic rehabilitation.
}

%%%%%%NEWW REFINED CONCLUSIONS%%%%%
\textcolor{black}{Future work will focus on expanding the evaluation of the proposed system to strengthen its clinical applicability. Specifically, the participant pool will be extended to include rehabilitation patients to assess the system’s effectiveness in clinical settings. A comparative study will evaluate differences between healthy subjects and patients, analyzing the impact of cognitive and psychological factors on rehabilitation performance.
In the scenario, the optimization of parameters ($m$, $b$, $\chi$, and $\delta$), whose effects on general human-robot interaction have been studied in this work, will address the specific needs of the therapeutic action to personalize the therapy for a given patient and a specific phase of the rehabilitative process.
}
%\textcolor{black}{Finally, further studies will focus on refining control strategies and optimizing key parameters ($m$, $b$, $\chi$, and $\delta$) to tailor the system’s behavior to different rehabilitation stages. These enhancements will ensure a more adaptive and clinically relevant robotic rehabilitation framework.}
\textcolor{black}{%Additionally, we aim to broaden the comparative analysis by including alternative control strategies and incorporating a wider range of evaluation metrics. This will allow for a more in-depth benchmarking of the system’s performance.
Another future direction involves integrating physiological metrics, such as electromyography (EMG), to complement subjective user evaluations and provide a more detailed assessment of patient effort and motor adaptation.
EMG signals will be directly integrated into the system to enhance the therapeutic action through an EMG-based assistance-as-needed mechanism.
}

\bibliographystyle{IEEEtran}
%\bibliography{Robotics4Rehabilitation}
\bibliography{MainVer11}
\begin{IEEEbiography}
[{\includegraphics[width=0.8in,height=1in,clip%,keepaspectratio
]{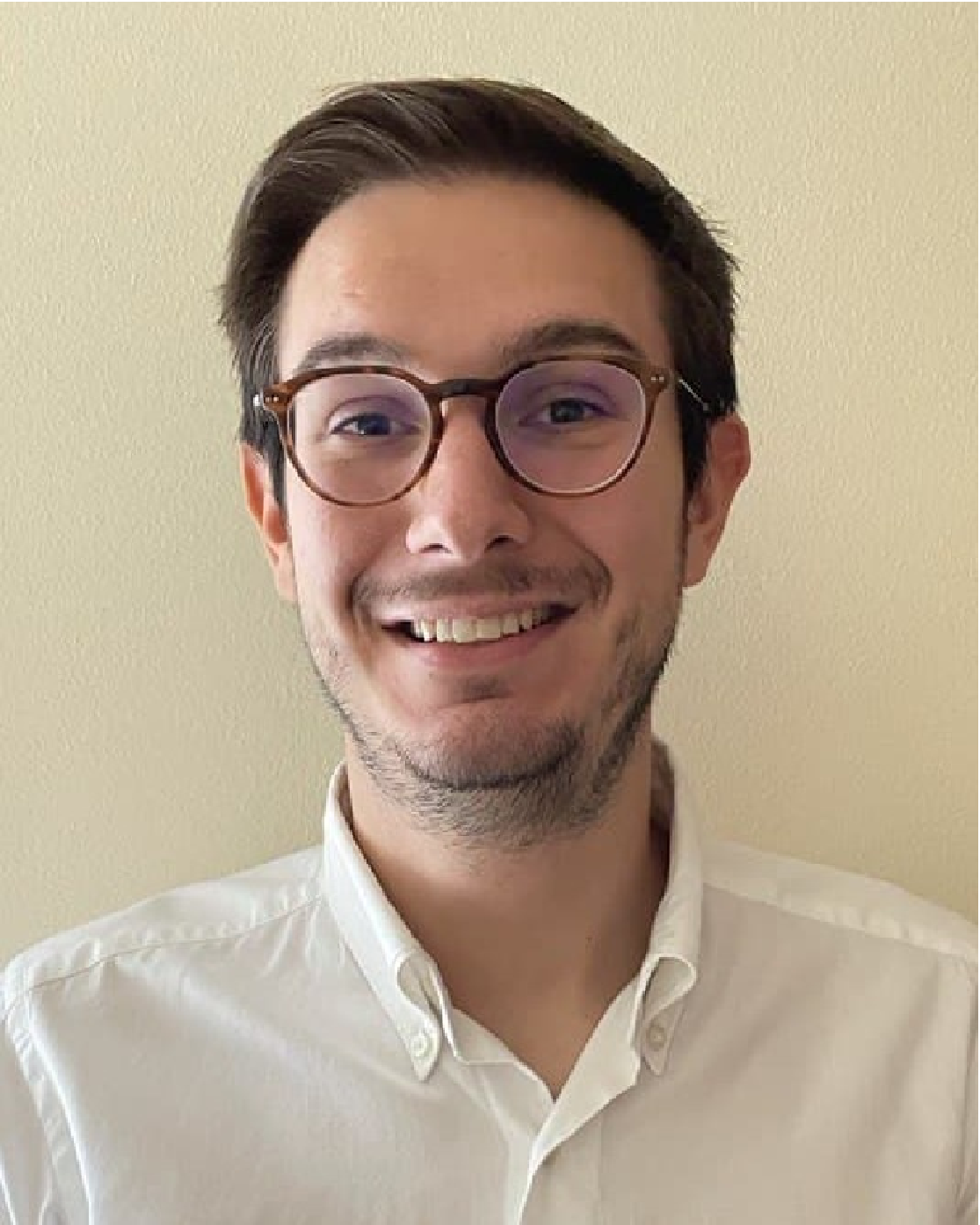}}]{Dario Onfiani}
received his B.Sc. and M.Sc. degrees in Mechanical Engineering from the University of Modena and Reggio Emilia, Italy, in 2019 and 2022, respectively. He is currently pursuing a Ph.D. degree in Information and Communication Technologies at the same university. %He is author and coauthor of two scientific papers presented at conferences.
His research interests include collaborative robotics, human-robot interaction, and the development of assistive robotic systems for rehabilitation.
\end{IEEEbiography}

\vskip -3.6\baselineskip plus -1fil
\begin{IEEEbiography}[{\includegraphics[width=0.8in,height=1in,clip%,keepaspectratio
]{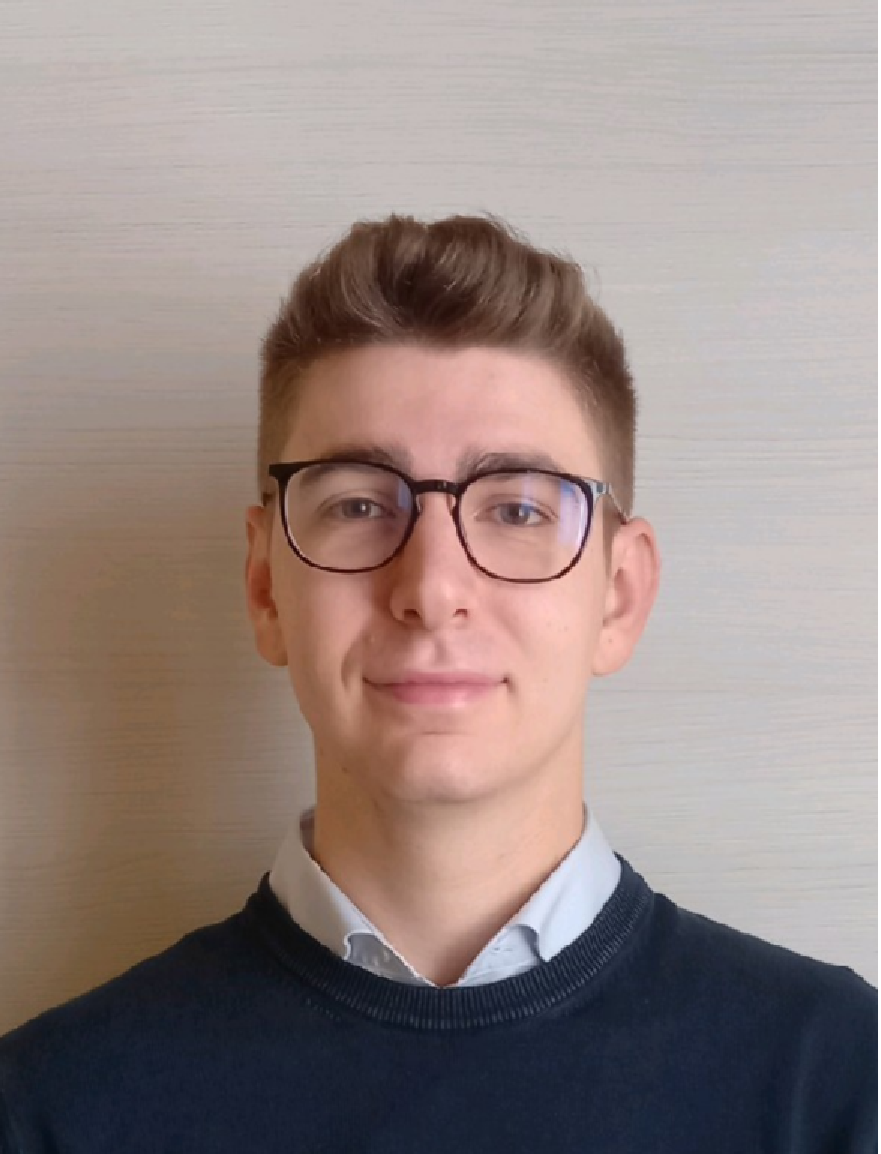}}]{Marco Caramaschi} received his B.Sc. and M.Sc. degrees in Mechanical Engineering from the University of Modena and Reggio Emilia, Italy, in 2019 and 2022, respectively. He worked on the development of assistive robotic systems for rehabilitation along his research fellowship at Department of Engineering ``Enzo Ferrari'' of the University of Modena and Reggio Emilia.
\end{IEEEbiography}

\vskip -2.9\baselineskip plus -1fil
\begin{IEEEbiography}[{\includegraphics[width=0.8in,height=1in,clip%,keepaspectratio
]{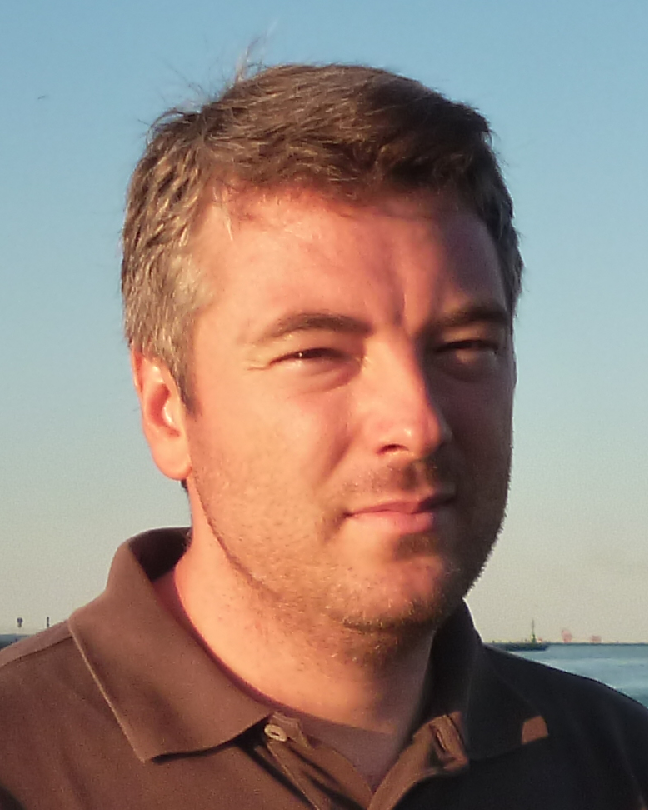}}]{Luigi Biagiotti}
 (Member, IEEE)  received the Ph.D. degree in automation engineering from the University of Bologna, Italy, in 2003. He is currently an Associate Professor with the Department of Engineering ``Enzo Ferrari'' of the University of Modena and Reggio Emilia.  %He is author or coauthor of more than 50 scientific papers presented at conferences or published in journals, and two books on motion planning and automatic control.
 His research interests include modelling and control of physical systems, trajectory planning and optimization,  control of robotic systems, human-robot interaction.
\end{IEEEbiography}

\vskip -2.9\baselineskip plus -1fil
\begin{IEEEbiography}[{\includegraphics[width=0.8in,height=1in,clip%,keepaspectratio
]{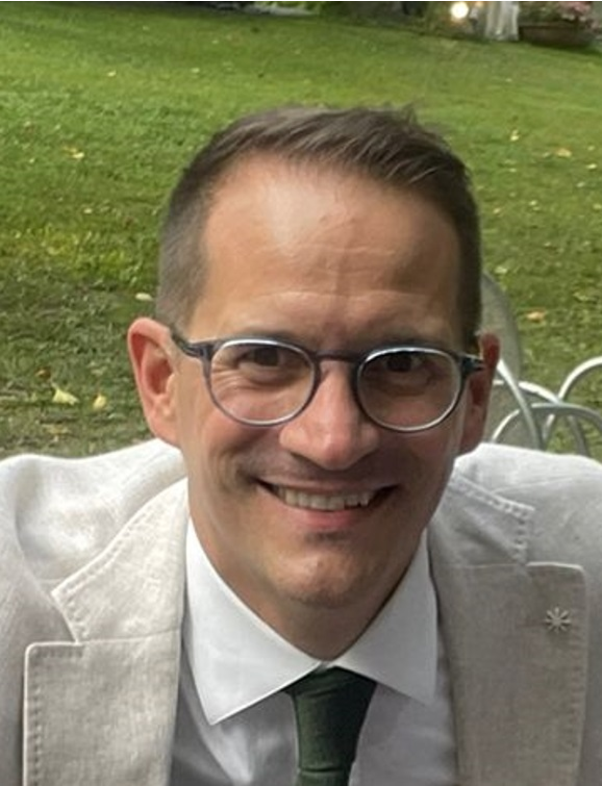}}]{Fabio Pini} received the Ph.D. degree in simulation methods and mechanical design from the University of Modena and Reggio Emilia, Italy, in 2008. He is currently a Associate Professor with the Department of Engineering ``Enzo Ferrari''.  His research interests include the definition of Computer Aided Design approaches for Product and Process Design, specifically oriented to the Development of Collaborative and Industrial robotics solutions and devices.
\end{IEEEbiography}

\end{document}